\documentclass[conference,letterpaper]{IEEEtran}
\usepackage[T1]{fontenc}
\usepackage{cite}
\usepackage{amsmath,amssymb}
\usepackage{graphicx}
\usepackage{booktabs,array,tabularx}
\usepackage{url}
\usepackage{balance}
\usepackage[hidelinks]{hyperref}
\graphicspath{{figures/}}
\newcommand{\TP}{\operatorname{TP}}
\newcommand{\Hent}{\operatorname{Ent}}
\newcommand{\R}{\mathbb{R}}
\newcommand{\Eset}{\mathcal{E}}

\begin{document}
\title{Topological Dynamics Under Non-Stationary Supervision:\\Persistent Homology of Neural Network Weight Evolution}
\author{\IEEEauthorblockN{Mitra Varun Gogulapati}
\IEEEauthorblockA{\textit{Independent Researcher}\\
mitravarun22@gmail.com}}
\maketitle

\begin{abstract}
We investigate how persistent topology of neuron-weight point clouds evolves under stable and non-stationary supervision. An identical multilayer perceptron is trained on Fashion-MNIST with correct labels, a fixed random permutation of the training-label vector, or a fresh permutation every epoch. Five paired seeds share initialization and minibatch order, yielding 15 models trained for 100 epochs. Raw and row-normalized incoming-weight clouds are analyzed with H0 and H1 Vietoris--Rips persistence at 22 checkpoints. Correct-label models reach $89.39\%$ mean true-label test accuracy; fixed-random models fit $67.94\%$ of their assigned training targets without generalizing to the original task. Correct-label models have descriptively greater time-averaged normalized fc3 H1 persistence than either random-label condition, but the configured exact comparisons do not reach conventional significance after Holm correction. Early condition classification achieves $60.0\%$ mean balanced accuracy using topology, $66.7\%$ using conventional diagnostics, and $60.0\%$ using both. Thus, no incremental predictive improvement is demonstrated in this setting. The archive contains all 3,300 configured topology cases. Interpretation is limited by five seeds, a single architecture and dataset, 282 capped H1 distance comparisons, and deferred bottleneck analysis.
\end{abstract}


\section{Introduction}
Ordinary supervised training assumes a fixed relationship between examples and targets. Relabeling, changing teachers, and online adaptation can instead modify the objective while optimization is in progress. Understanding such dynamics requires separating target informativeness from target stability. A correct-versus-changing-label comparison alone cannot distinguish the two: random targets remove useful image--label structure, and changing them adds non-stationarity.

Random-label memorization demonstrates that neural networks can fit mappings unrelated to input semantics, although its extent depends on model, optimization, and training budget~\cite{zhang2017,arpit2017}. Work on plasticity further establishes that networks can become less able to adapt during continued learning~\cite{lyle2023,dohare2024}. Our changing-random condition is an extreme supervision intervention, not a standard continual-learning benchmark; we do not measure later-task plasticity or retention. A fixed-random control is essential because it retains an arbitrary but temporally stable assignment.

Loss and accuracy quantify fitting and generalization, but do not exhaustively describe internal organization. Persistent homology~\cite{edelsbrunner2002} supplies a multiscale summary of a chosen representation; persistence landscapes provide vector-valued summaries~\cite{bubenik2015}. Applied to incoming-weight vectors, these tools may expose changes in neuron similarity. Equally, the summaries may recapitulate ordinary geometric or spectral changes. Demonstrating a visually different trajectory is therefore weaker than demonstrating useful information beyond a conventional baseline.

We ask four questions: \emph{RQ1}, how do topological trajectories differ across correct, fixed-random, and changing-random supervision? \emph{RQ2}, how do raw and normalized representations affect the comparison? \emph{RQ3}, how do topological changes associate with optimization diagnostics? \emph{RQ4}, do early topological features improve condition prediction over simpler measurements?

Our contributions are a five-seed paired benchmark, a reproducible 22-checkpoint analysis of five hidden layers, a complete accounting of numerical coverage, and a grouped evaluation of incremental predictive value. The empirical scope is intentionally narrow. The results include partial arbitrary-label fitting, descriptive persistence differences, limited exact-test resolution, and no demonstrated classification gain from adding topology. Neither causality nor a topological phase transition is claimed.

\begin{figure*}[!t]
\centering
\includegraphics[width=\textwidth]{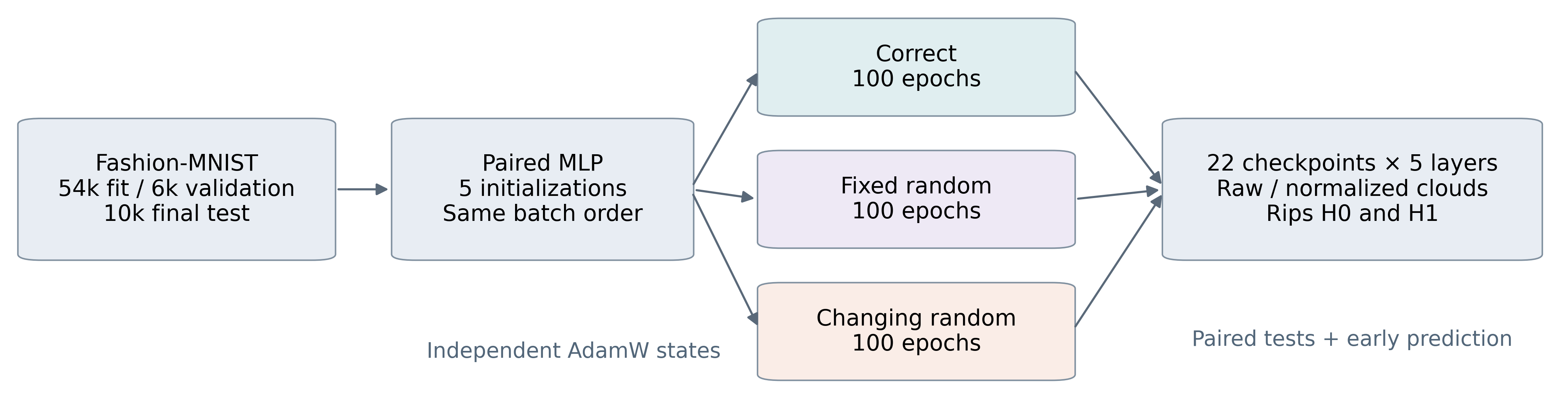}
\caption{Experimental design. A shared fitting/validation partition and paired initialization/order support three independent supervision runs per seed. The topology is computed on incoming-weight row clouds at 22 checkpoints, not on the complete parameter space or the loss landscape. The official test set is reserved for final evaluation.}
\label{fig:pipeline}
\end{figure*}

\section{Related Work}
\subsection{What Object Is Being Analyzed?}
Neural persistence~\cite{rieck2019} analyzes a weighted network graph using zero-dimensional persistence. Convolutional persistence studies topological descriptors of network weights and filters~\cite{khramtsova2023}. These precedents mean that computing persistence during training is not itself a new contribution. Our object is a Euclidean cloud whose points are complete incoming-weight vectors. Its filtration values are distances between rows, rather than individual connection magnitudes. In particular, H1 loops in this Rips complex are not literal cycles in the feed-forward network.

Girrbach \emph{et al.}~\cite{girrbach2023} identify substantial weight-variance dependence in a graph-based neural-persistence measure. Their result does not establish the same limitation for our construction, but motivates explicit scale controls and baseline comparisons. Activation-based analyses, including topology of data representations across layers~\cite{naitzat2020}, study another object that depends on probe inputs. We do not measure activation topology here.

\subsection{Supervision, Geometry, and Evaluation}
The random-label literature~\cite{zhang2017,arpit2017} motivates an arbitrary-assignment control, while plasticity studies~\cite{lyle2023,dohare2024} motivate studying changing targets without equating that intervention with loss of plasticity. Spectral analyses~\cite{mahoney2019} motivate our singular-value and effective-rank baselines. Function-preserving reparameterizations can alter parameter geometry~\cite{dinh2017}; row normalization should therefore not be mistaken for invariance to all equivalent network parameterizations.

The contribution is the controlled three-condition comparison and its baseline evaluation, rather than a claim of a first general connection between topology and learning. We adopt seed-level replication and artifact transparency in the spirit of reproducibility recommendations~\cite{pineau2021}. Multiple epochs or intervals from one run do not create additional independent replications.

\section{Experimental Methods}
\subsection{Data, Network, and Optimization}
Fashion-MNIST~\cite{xiao2017} contains ten classes of $28\times28$ grayscale images. A stratified split of the official training set produces 54,000 fitting and 6,000 validation examples, shared by all conditions and seeds. The partition seed is 314159. Fitting-image pixel mean and population standard deviation, after division by 255, normalize all splits. Validation images never enter fitting or label permutation. Images are flattened without augmentation. The official 10,000-image test set is evaluated at the end of each completed 100-epoch run and is not used for feature or hyperparameter selection.

The architecture is
\begin{equation}
784\!\rightarrow\!1024\!\rightarrow\!512\!\rightarrow\!256\!\rightarrow\!128\!\rightarrow\!64\!\rightarrow\!10.
\end{equation}
The five hidden linear layers, \texttt{fc1}--\texttt{fc5}, use ReLU; the \texttt{output} layer supplies raw logits. There is no dropout or batch normalization. Biases are trained and saved, but are not coordinates in the weight clouds. The implementation uses PyTorch~\cite{paszke2019} and AdamW~\cite{loshchilov2019}. Table~\ref{tab:settings} gives the fixed training configuration.

\begin{table}[!t]
\centering
\caption{Dataset and training configuration.}
\label{tab:settings}
\footnotesize
\begin{tabularx}{\columnwidth}{@{}lX@{}}
\toprule
Item & Setting \\
\midrule
Fitting / validation / test & 54,000 / 6,000 / 10,000 \\
Optimizer / loss & AdamW / cross-entropy \\
Learning rate / weight decay & Constant 0.001 / 0.01 \\
Batch size / epochs & 256 / 100 \\
Gradient clipping & Global L2 limit 1.0 \\
Paired seeds & 42, 123, 456, 789, 2026 \\
Conditions / trained models & Three / 15 \\
Topology checkpoints & 22, including initialization \\
Cloud size & 128 in fc1--fc4; 64 in fc5 \\
Representations / homology & Raw and normalized / H0 and H1 \\
Compute & One NVIDIA Tesla T4; CPU topology \\
\bottomrule
\end{tabularx}
\end{table}

Online loss is the sample-weighted minibatch loss encountered during the epoch. Gradient norm is the sample-weighted mean of global minibatch L2 norms measured before clipping. Weight norm is the global L2 norm of all trainable parameters, including biases. Update norm is the global displacement between consecutive epoch parameter states. Assigned-target training accuracy, original-label training accuracy, and original-label validation accuracy use the end-of-epoch model. These quantities answer different questions and are not conflated.

\subsection{Paired Conditions and Randomness}
For each seed, one initialization is cloned into three models with independent AdamW states. Minibatch order is a deterministic function of seed and epoch and is identical across conditions. Model initialization, sample shuffling, and target assignment use separate random streams.

In the \emph{correct} condition, the original fitting labels remain fixed. In \emph{fixed random}, one permutation of the fitting-label vector is reused for all epochs. In \emph{changing random}, a new permutation is generated at each epoch boundary and retained throughout that epoch. This shuffles labels \emph{across examples}; it is not a permutation of the ten class identities. Each assignment preserves the exact class histogram, and a SHA-256 hash is logged. The archive confirms a single target hash per correct/fixed run and 100 distinct hashes per changing run. The fixed condition is a stable arbitrary-target control, not an assumption of successful full memorization.

\subsection{Point-Cloud Construction}
For $W_\ell\in\R^{N_\ell\times D_\ell}$, a neuron contributes the row vector $w_i\in\R^{D_\ell}$. Fixed sampled indices $I_\ell$ define
\begin{equation}
\begin{aligned}
 X_\ell^{\rm raw}&=\{w_i:i\in I_\ell\},\\
 X_\ell^{\rm norm}&=\left\{\frac{w_i}{\max(\|w_i\|_2,10^{-12})}:i\in I_\ell\right\}.
\end{aligned}
\label{eq:clouds}
\end{equation}
Indices are sampled without replacement using a NumPy generator seeded by $2718+N_\ell$ and sorted. They are reused across all seeds, conditions, epochs, and representations. All 64 fc5 rows are retained. An audit of all 3,300 saved clouds found no zero-norm sampled rows, identical per-layer index subsets, and identical initialization clouds across paired conditions.

Raw clouds mix directional and radial changes; normalized clouds remove individual row magnitudes. Neither normalization nor diagram comparison removes all architectural, sampling, or parameterization dependence. Layers have different ambient dimensions and point counts. We do not pool them into a single cloud or interpret their scales as automatically equivalent.

\subsection{Persistence and Numerical Summaries}
Ripser.py computes Vietoris--Rips persistence~\cite{bauer2021} from the Euclidean distance matrix with maximum homology dimension 1, default $\mathbb F_2$ coefficients, and no finite filtration truncation. H2 is not part of the primary computational budget. Stability theory motivates persistence as a perturbation-aware representation~\cite{cohen2007}, but does not by itself imply that a particular summary measures generalization or is invariant to network reparameterization.

For finite intervals $(b_j,d_j)$ with lifetime $a_j=d_j-b_j$, total persistence and entropy are
\begin{equation}
\begin{aligned}
 \TP(D)&=\sum_j a_j,\qquad p_j=\frac{a_j}{\TP(D)},\\
 \Hent(D)&=-\sum_{j:a_j>0}p_j\log p_j.
\end{aligned}
\label{eq:summary}
\end{equation}
Empty diagrams or zero total persistence give entropy zero. We also save maximum/mean finite lifetime and positive-lifetime feature count. The single essential H0 interval is omitted from finite-lifetime summaries and matching distances, but retained in the H0 Betti curve. Betti counts are $\beta_k(r)=\#\{j:b_j\leq r<d_j\}$; a diagram's total feature count is not its Betti number at every scale. Landscapes~\cite{bubenik2015} are the ordered persistence tent functions.

The exact checkpoint set is
\begin{equation}
\begin{split}
\Eset={}&\{0,1,\ldots,10,15,20,25,30,40,50,60,\!\!\\[-2pt]
&70,80,90,100\},\qquad |\Eset|=22.
\end{split}
\end{equation}
The recorded protocol note saying 23 checkpoints is a counting error; the configured schedule and all saved topology cases agree at 22. Betti curves use 128 grid values and landscapes use 32 values and three ranks. Normalized grids are fixed on $[0,2]$; each raw layer uses $[0,1.001\max d_{ij}]$ over all analyzed cases. Appendix~\ref{app:definitions} specifies which grid-dependent features may enter early prediction.

\subsection{Diagram Distances and Change Rates}
Persim implements a finite-diagram 1-Wasserstein matching sum with Euclidean ground cost:
\begin{equation}
 W_{1,2}(D,D')=\min_{\gamma}\sum_{u\in\widetilde D}\|u-\gamma(u)\|_2,
\label{eq:wasserstein}
\end{equation}
where the diagrams are augmented by diagonal points to permit unmatched intervals. The diagonal cost of $(b,d)$ is $(d-b)/\sqrt{2}$. Empty--empty distance is zero; an empty--nonempty distance is the sum of these diagonal costs. This is a matching \emph{sum}, not a feature-count-normalized mean.

We compare successive sampled checkpoints, matched conditions within a seed, and seeds within a condition. A temporal rate is $r_k=W_{1,2}(D_{e_{k-1}},D_{e_k})/(e_k-e_{k-1})$. It is an average change over the sampled interval, not a measurement at every intervening epoch. The saved evaluation contains 26,100 comparisons across homology dimensions, layers, representations, and comparison types. A comparison is skipped if either finite diagram exceeds 400 features. All 282 skips occur in normalized H1; they are missing values, never zeros. Bottleneck computations were deferred and no bottleneck values are reported.

\subsection{Optimization Associations and Early Prediction}
The exploratory association analysis first divides adjacent checkpoint differences by elapsed epochs, then computes within-seed Spearman correlations between topological-summary rates and loss, gradient norm, update norm, validation accuracy, or weight norm. The saved topological summaries are H0 total persistence, H1 total persistence, and H1 entropy. Lag 1 shifts the topological-rate sequence by one \emph{sampled checkpoint}; it is not a fixed one-epoch lag. These are not correlations of Wasserstein rate unless explicitly computed separately. No causal effect is estimated.

For early condition classification, one example per seed/condition is constructed by averaging features from epochs 1--5. The topology set contains finite H0/H1 summaries, normalized Betti areas and landscapes, and available temporal Wasserstein rates. Raw landscapes and raw Betti areas are excluded because their grid range uses later checkpoints. Normalized grids are fixed independently of later training. The baseline contains sampled-cloud row-norm and pairwise-distance statistics, three leading singular values and effective rank, plus loss, gradient norm, global weight norm, and update norm. Detailed feature accounting appears in Appendix~\ref{app:prediction}.

Scikit-learn~\cite{pedregosa2011} pipelines perform median imputation, standardization, and L2-regularized multinomial logistic regression. Nested leave-one-seed-out evaluation holds out all three models from one seed. The other four seeds supply 12 training examples; inner seed-held-out folds select $C\in\{0.01,0.1,1,10\}$. All preprocessing is fitted inside the corresponding training fold. Outer balanced accuracy is the primary metric. Since every test seed has exactly one example per condition, a fold score is a multiple of $1/3$. The outer training sets overlap, so comparisons across outer folds are descriptive rather than independent test replications.

\section{Seed-Level Statistical Analysis}
The configured primary representation is normalized fc3, not a layer selected by searching the presented plots. For seed $s$ and condition $c$, the temporal-average endpoint is
\begin{equation}
 A_{s,c}=\frac{1}{100}\sum_{k=1}^{21}\frac{\TP_{s,c,e_{k-1}}+\TP_{s,c,e_k}}{2}(e_k-e_{k-1}).
\label{eq:average}
\end{equation}
This is trapezoidal integration over the irregular schedule, not an unweighted average of 22 checkpoints. The second endpoint is the arithmetic mean of $r_k$ over available normalized fc3 H1 temporal comparisons; unavailable intervals are omitted, making that endpoint conditional on coverage. The third comparison is topology minus baseline outer-fold balanced accuracy.

For each comparison, $\delta_s$ denotes a paired seed difference. We report $\bar\delta$, sample SD, $d_z=\bar\delta/s_\delta$, and an unadjusted $t_4$ interval $\bar\delta\pm t_{4,0.975}s_\delta/\sqrt5$. The exact two-sided sign-flip test enumerates all $2^5$ sign patterns and compares absolute mean differences. Its distributional interpretation assumes sign symmetry under the null; seeds, not epochs or intervals, are the units. The minimum attainable two-sided $p$ is $2/32=0.0625$, preventing a conventional $0.05$ rejection even before multiplicity adjustment.

Holm correction~\cite{holm1979} includes all seven configured comparisons: three condition contrasts for each of the two topology endpoints and one topology-versus-baseline classification contrast. There is no claim of external preregistration. Table~\ref{tab:tests} reports the full family, including negative comparisons. Classification-fold tests are retained for transparency but are not assigned independent-replication status. The later visualization intervals, PCA, clustering, and association networks are exploratory. Different assumptions underlie an unadjusted $t$ interval and an exact sign-flip test; exclusion of zero by the former is not a corrected significance result.

\section{Results}
\subsection{Coverage and Final Performance}
All 15 configured models have contiguous 100-epoch histories and final test records. All $15\times22\times5\times2=3300$ topology cases are present. The 26,100 comparison records comprise 6,300 temporal, 6,600 matched-condition, and 13,200 across-seed evaluations. Of these, 25,818 have finite Wasserstein values. Appendices~\ref{app:coverage} and~\ref{app:execution} distinguish complete case coverage from the deliberate distance skips.

\begin{table*}[!t]
\centering
\caption{Final performance across five paired seeds (percent).}
\label{tab:performance}
\footnotesize
\setlength{\tabcolsep}{4pt}
\begin{tabular}{lrrr}
\toprule
Condition & Assigned train & True validation & True test \\
\midrule
Correct & $99.51 \pm 0.21$ & $89.78 \pm 0.19$ & $89.39 \pm 0.24$ \\
Fixed random & $67.94 \pm 4.89$ & $10.10 \pm 0.40$ & $10.29 \pm 0.61$ \\
Changing random & $10.00 \pm 0.00$ & $10.00 \pm 0.00$ & $10.00 \pm 0.00$ \\
\bottomrule
\end{tabular}
\par\vspace{2pt}\begin{minipage}{\textwidth}\scriptsize Values are means $\pm$ one sample standard deviation across seeds. Assigned-target accuracy is evaluated using end-of-epoch weights. Test labels remain correct in every condition.\end{minipage}
\end{table*}

\begin{figure*}[!t]
\centering\includegraphics[width=\textwidth]{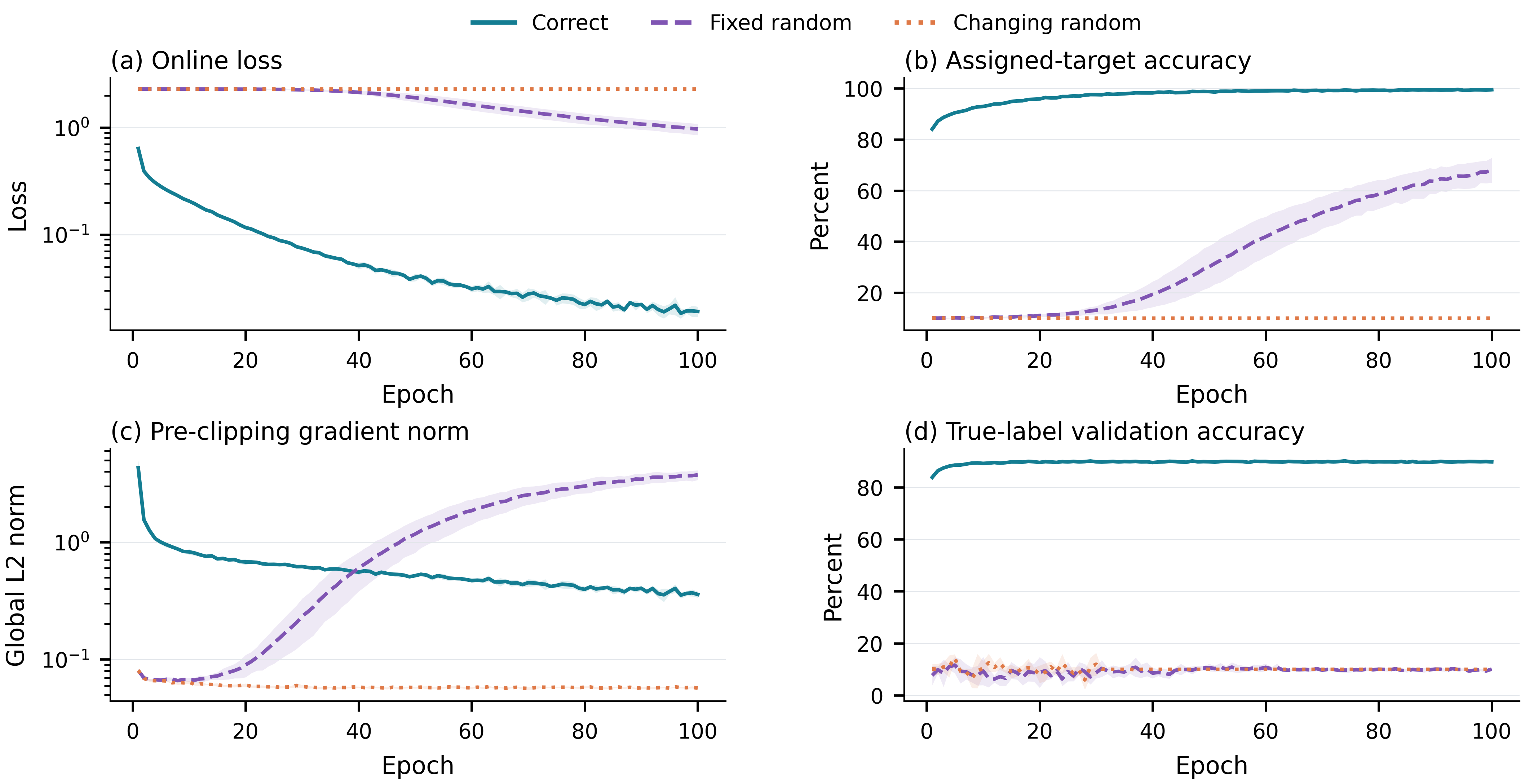}
\caption{Training behavior across five seeds. Lines are condition means; bands are $\pm1$ sample SD across seeds, not confidence intervals. Correct-label models fit and generalize; fixed-random models progressively fit their assigned targets while remaining near chance on the original task; changing-random models remain near chance under the recorded configuration. Online loss and end-of-epoch accuracy have different measurement conventions.}
\label{fig:training}
\end{figure*}

Correct-label models achieve $99.51\%$ assigned-target training accuracy and $89.39\%$ true-label test accuracy on average (Table~\ref{tab:performance}). Fixed-random models achieve $67.94\%$ against assigned training targets and $10.29\%$ on the true test task. Thus the fixed-random control exhibits partial fitting, not complete arbitrary-label memorization. Changing-random models record $10.00\%$ assigned-target and true-label test accuracy. This is an observed outcome of this schedule and optimizer, not a proof that learning within an epoch of changing targets is impossible. Figure~\ref{fig:training} shows the accompanying optimization trajectories.

\subsection{Topological Trajectories and Layer Dependence}
\begin{figure*}[!t]
\centering\includegraphics[width=\textwidth]{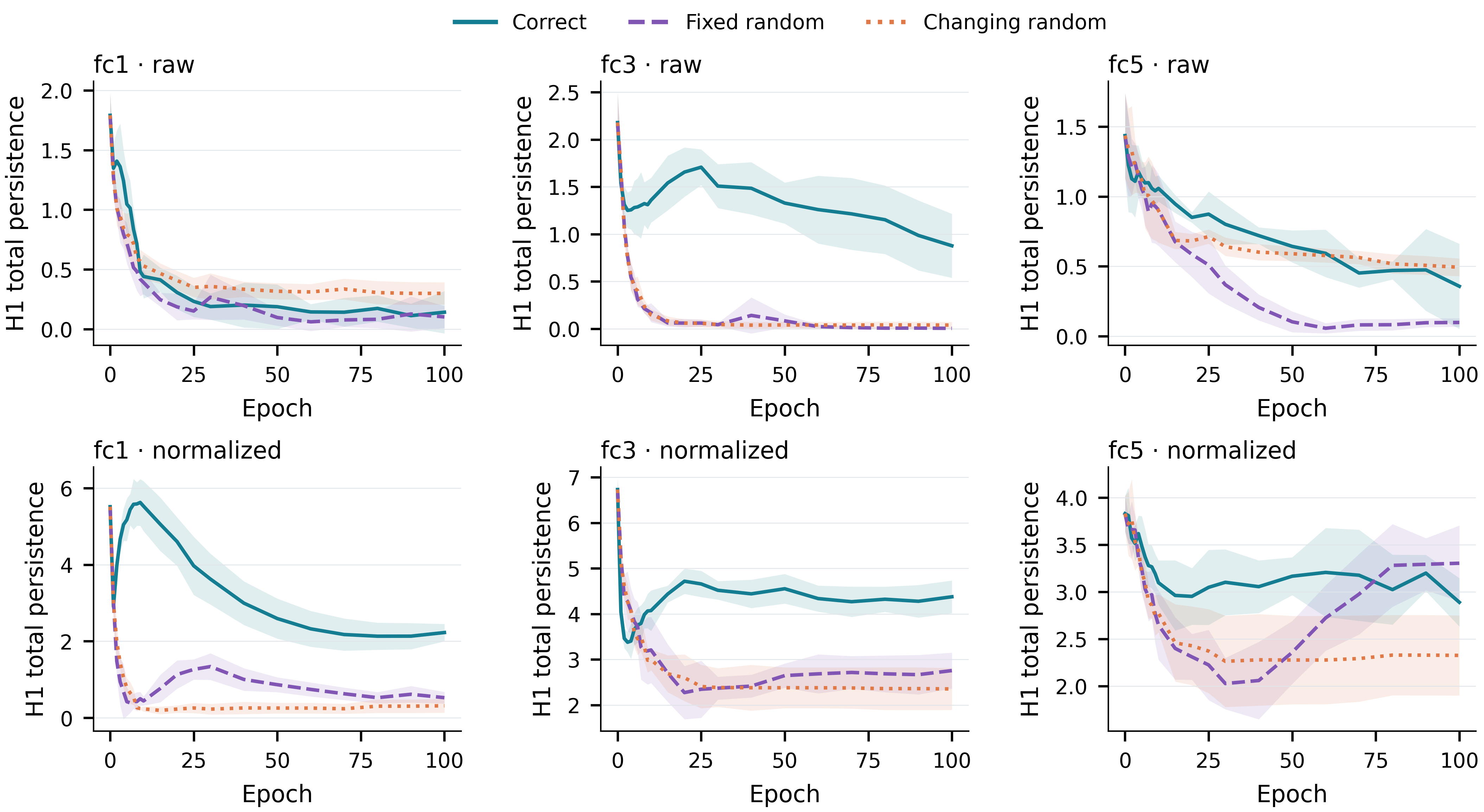}
\caption{H1 total-persistence trajectories in representative early, middle, and late layers. Raw (top) and normalized (bottom) representations are kept separate. Means and $\pm1$ seed SD are shown; all 22 configured checkpoints are present. Connecting lines are guides between observations, not additional measured checkpoints. Different ambient dimensions and the 64-point fc5 cloud limit direct cross-layer interpretation.}
\label{fig:topology}
\end{figure*}

Figure~\ref{fig:topology} separates raw and normalized H1 trajectories. Raw curves mix magnitude and direction; normalization changes the relative appearance of conditions and layers. There is no single universal trajectory ordering. For example, final normalized H1 persistence can differ substantially between early and late layers, and fc5 uses half as many points as fc1--fc4. Complete final-checkpoint summaries appear in Table~\ref{tab:topofinal}.

For the configured normalized fc3 endpoint, correct-label temporal-average H1 persistence exceeds fixed-random by $1.6156$ and changing-random by $1.7836$ units. The signs agree in all five paired seeds for both contrasts. However, each raw exact $p$ is $0.0625$ and each Holm-adjusted $p$ is $0.4375$. Fixed-random minus changing-random is $0.1680$ units with adjusted $p=1.0000$. These numerical differences are descriptive, not conventional significance claims.

\begin{figure*}[!t]
\centering\includegraphics[width=\textwidth]{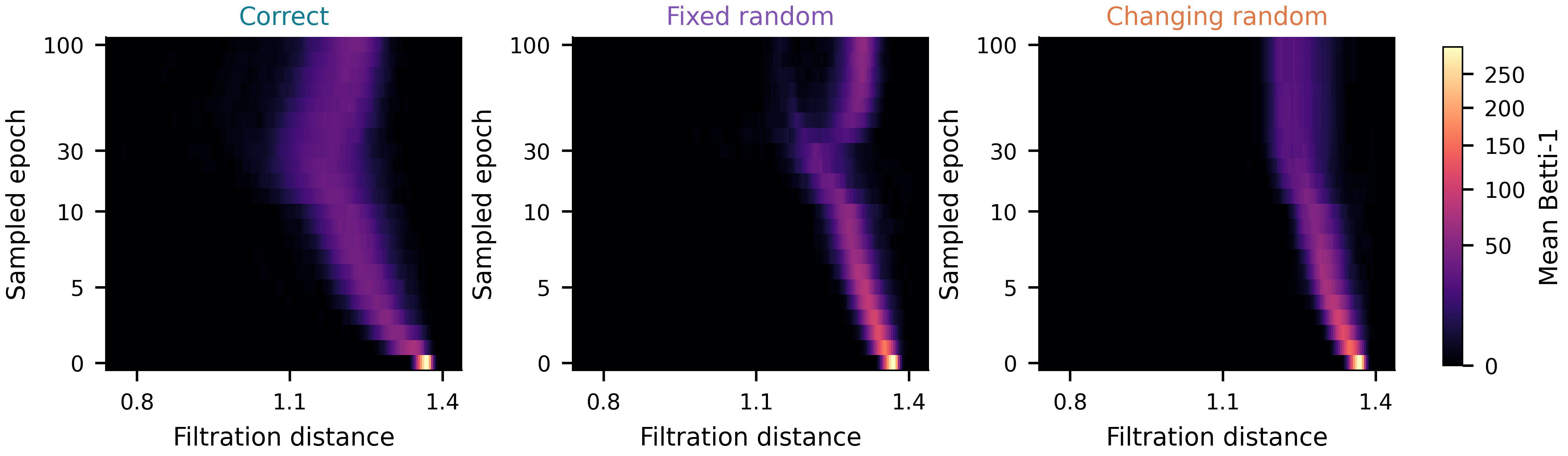}
\caption{Normalized fc3 H1 Betti surfaces reconstructed from saved finite intervals and averaged across five seeds. Columns use the same filtration window and nonlinear color normalization ($\gamma=0.55$) to reveal low-count structure without changing the counts. Vertical positions enumerate the 22 observed checkpoints equally, not elapsed training time. The 160-value display grid is separate from the 128-value analysis grid. Betti counts are not total persistent-feature counts.}
\label{fig:betti}
\end{figure*}

Normalized H0 total persistence falls from initialization to epoch 100 in every layer/condition mean; the final magnitudes differ across conditions. This is a finite-interval merging-scale summary, not the number of components at every scale. Figure~\ref{fig:betti} instead displays actual live H1 counts as a function of filtration. Its concentration and scale shifts explain why one scalar total-persistence summary does not specify the full diagram geometry.

\begin{table*}[!t]
\centering
\caption{Complete configured seven-comparison family; $n=5$ seed-level differences.}
\label{tab:tests}
\footnotesize
\setlength{\tabcolsep}{4pt}
\begin{tabular}{llrrrrr}
\toprule
Outcome & Comparison & Mean $\Delta$ & Unadjusted 95\% $t$ interval & $d_z$ & Exact $p$ & Holm $p$ \\
\midrule
$\overline{\mathrm{TP}}_1$ & Fixed random $-$ Changing random & 0.1680 & $[-0.3488, 0.6849]$ & 0.404 & 0.5000 & 1.0000 \\
$\overline{\mathrm{TP}}_1$ & Correct $-$ Fixed random & 1.6156 & $[1.3087, 1.9224]$ & 6.538 & 0.0625 & 0.4375 \\
$\overline{\mathrm{TP}}_1$ & Correct $-$ Changing random & 1.7836 & $[1.0857, 2.4815]$ & 3.173 & 0.0625 & 0.4375 \\
$\overline{r}_1$ & Fixed random $-$ Changing random & 0.0952 & $[-0.0453, 0.2358]$ & 0.841 & 0.1875 & 0.9375 \\
$\overline{r}_1$ & Correct $-$ Fixed random & -0.0407 & $[-0.1069, 0.0256]$ & -0.762 & 0.1875 & 0.9375 \\
$\overline{r}_1$ & Correct $-$ Changing random & 0.0546 & $[-0.0541, 0.1632]$ & 0.624 & 0.2500 & 0.9375 \\
$\Delta\mathrm{BA}$ & Topology $-$ Baseline & -6.6667 & $[-41.2950, 27.9617]$ & -0.239 & 1.0000 & 1.0000 \\
\bottomrule
\end{tabular}
\par\vspace{2pt}\begin{minipage}{\textwidth}\scriptsize $\overline{\mathrm{TP}}_1$: time-averaged normalized fc3 H1 total persistence; $\overline r_1$: mean available temporal Wasserstein rate; $\Delta\mathrm{BA}$: topology minus baseline, in percentage points. $d_z$ is the mean difference divided by its sample SD. Classification-fold tests are descriptive because outer training sets overlap.\end{minipage}
\end{table*}

\begin{figure*}[!t]
\centering\includegraphics[width=\textwidth]{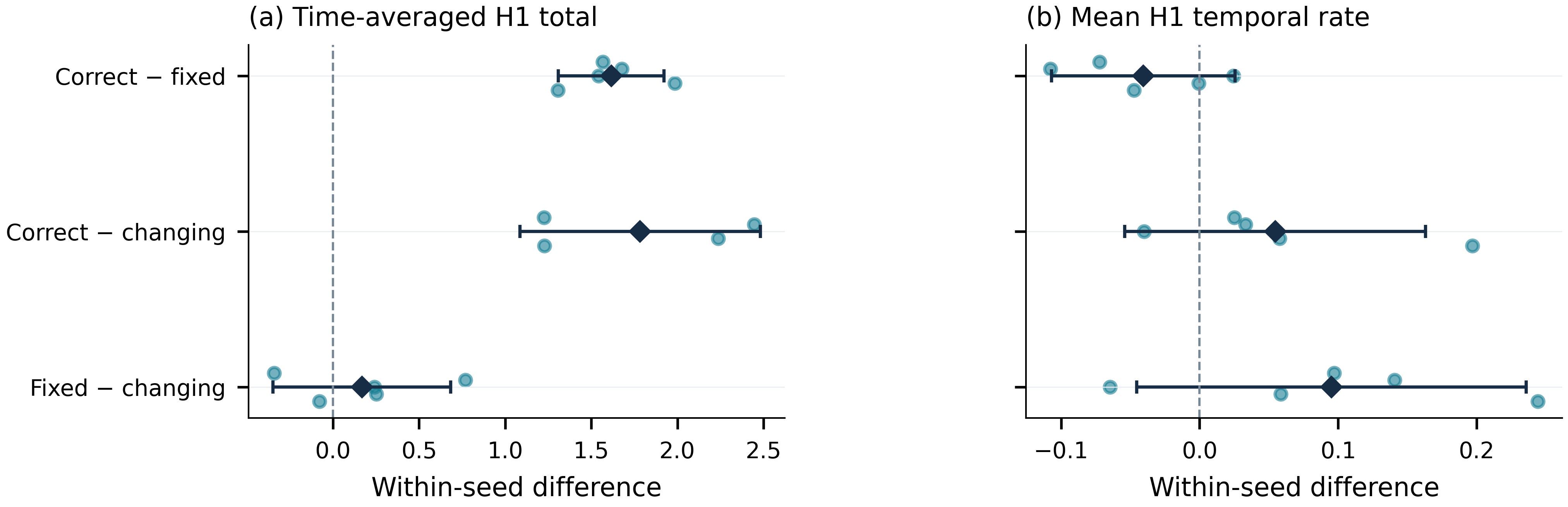}
\caption{Configured normalized fc3 contrasts. Small points are five individual paired seed differences; diamonds and bars show the mean and unadjusted 95\% $t_4$ interval. Endpoint (a) is time-averaged H1 total persistence; (b) is mean available temporal Wasserstein rate. These intervals are not multiplicity-corrected and do not replace the exact/Holm results in Table~\ref{tab:tests}.}
\label{fig:effects}
\end{figure*}

\subsection{Distances and Optimization Associations}
Mean temporal-rate differences are smaller relative to seed variability than the total-persistence contrasts (Fig.~\ref{fig:effects}). Fixed-random minus changing-random is $0.0952$, correct minus fixed-random is $-0.0407$, and correct minus changing-random is $0.0546$; adjusted $p$ values are $0.9375$ in all three cases. No claim that moving targets necessarily produce greater diagram displacement is supported by these comparisons.

All 282 capped distances are normalized H1 cases in fc1--fc4. Of these, 180 are recorded at epoch 0, 95 at epoch 1, and seven at epoch 2; temporal rows are indexed by their later endpoint. Missingness is therefore concentrated near initialization, not spread uniformly across training. This matters especially for early-change-rate features and comparisons conditional on available diagrams. H0 and raw H1 distances are fully available. Exact breakdowns are given in Tables~\ref{tab:skips} and~\ref{tab:skipepoch}.

The archived association analysis contains 4,500 within-seed feature/metric/lag correlations. Table~\ref{tab:corr} reports the normalized fc3 H1 subset; remaining measurements are supplied as CSV. Changes are normalized by elapsed epochs, but common trends or causes may remain. We do not interpret the exploratory association networks as causal circuits or use their visualization thresholds as tests.

\subsection{Early Classification and Incremental Value}
\begin{figure*}[!t]
\centering\includegraphics[width=\textwidth]{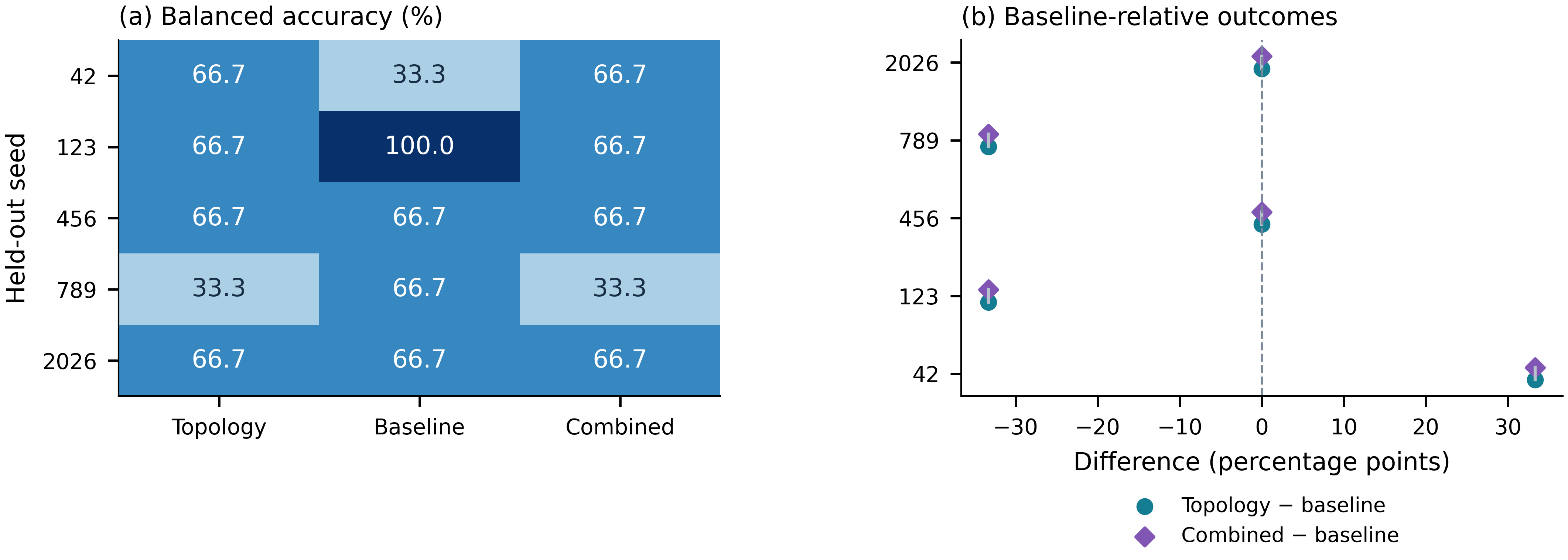}
\caption{Early condition classification with nested seed-held-out evaluation. (a) Actual fold-level balanced accuracies. (b) Paired baseline-relative differences in percentage points; overlapping topology/combined outcomes are offset vertically for visibility. Each fold has only three test examples, and outer training sets overlap. The plot retains the negative incremental-value result rather than implying that condition separation in full-trajectory figures establishes predictive usefulness.}
\label{fig:classification}
\end{figure*}

Table~\ref{tab:prediction} and Fig.~\ref{fig:classification} show every outer fold. Topology achieves $60.0\%$ mean balanced accuracy, the baseline $66.7\%$, and the combined set $60.0\%$. Thus the tested topology features do not demonstrate improvement beyond the conventional diagnostics. The difference is one additional correct assignment among 15 held-out models, not compelling evidence that either family is generally superior. No confusion matrix is inferred because individual predicted labels were not retained.

\begin{table}[!t]
\centering
\caption{Nested leave-one-seed-out balanced accuracy (percent).}
\label{tab:prediction}
\footnotesize
\setlength{\tabcolsep}{4pt}
\begin{tabular}{lrrr}
\toprule
Held-out seed & Topology & Baseline & Combined \\
\midrule
42 & 66.67 & 33.33 & 66.67 \\
123 & 66.67 & 100.00 & 66.67 \\
456 & 66.67 & 66.67 & 66.67 \\
789 & 33.33 & 66.67 & 33.33 \\
2026 & 66.67 & 66.67 & 66.67 \\
Mean & 60.00 & 66.67 & 60.00 \\
\bottomrule
\end{tabular}
\par\vspace{2pt}\begin{minipage}{\columnwidth}\scriptsize Each held-out seed contains three models, one per condition. The 6.67-point mean difference equals one additional correct assignment among 15 held-out models; individual predicted labels were not retained.\end{minipage}
\end{table}

The nominal input dimensionalities are 1,090 topology features, 94 baseline features, and 1,184 combined features, compared with 12 examples in an outer training fold. Regularization and nested selection limit direct leakage, but do not resolve this severe small-sample/high-dimensional setting or unequal feature budgets. The original descriptive topology-minus-baseline comparison is $-6.67$ percentage points, with exact and adjusted $p=1.0000$; its broad unadjusted interval is reported in Table~\ref{tab:tests}.

\subsection{Expanded Descriptive Evidence}
The appendix separates optimization, representation, and multivariate views rather than relying on a single topology trajectory. Tables~\ref{tab:milestones} and~\ref{tab:optimization} give five training milestones for every condition, with seed variability retained. Table~\ref{tab:layer_averages} extends the time-weighted H1 summary across all five hidden layers; Table~\ref{tab:final_geometry} pairs endpoint topology with geometric and spectral measurements. These additional summaries are exploratory and do not expand the primary multiplicity-adjusted testing family.

Quantitatively, normalized fc3 time-averaged H1 total persistence is $4.359\pm0.204$ under correct labels, $2.743\pm0.227$ under fixed random labels, and $2.575\pm0.408$ under changing random labels (mean $\pm$ sample SD across seeds). Raw fc3 values are $1.314\pm0.175$, $0.110\pm0.031$, and $0.110\pm0.028$, respectively. Representation therefore changes both scale and the visible separation between randomized conditions. Depth also matters: normalized fc4 averages are $5.215\pm0.145$, $5.134\pm0.540$, and $4.115\pm0.378$, so the correct-versus-fixed separation is much smaller there than at fc3. These descriptive observations support reporting the full layer profile rather than generalizing one layer's ordering to the entire network.

At epoch 50, fixed-random assigned-target accuracy is $30.127\pm8.184\%$ and rises to $67.938\pm4.887\%$ by epoch 100, while true-label validation accuracy remains near chance. Correct-label validation accuracy is $89.830\pm0.274\%$ at epoch 50 and $89.783\pm0.187\%$ at epoch 100, even as assigned-target accuracy rises from $98.766\pm0.098\%$ to $99.506\pm0.206\%$. Thus additional fitting over this interval does not produce an accompanying increase in mean validation accuracy. The endpoint true-label training--validation gap is $9.722\pm0.175$ percentage points for correct supervision. These are measured trajectory summaries, not a separately tested claim about overfitting onset.

The distinction between scale and lifetime distribution is visible in the all-layer H0 and H1-entropy profiles (Figs.~\ref{fig:h0all} and~\ref{fig:entropyall}). The birth--lifetime atlas (Fig.~\ref{fig:diagramatlas}) exposes the finite intervals behind aggregate values, while geometry--topology trajectories (Fig.~\ref{fig:phaseatlas}) show how persistence relates to pairwise cloud separation. Neither a visually separated curve nor a dense interval cloud establishes a statistically supported supervision effect beyond the primary analysis.

The retrospective clustering, change-rate association atlas, and temporal summary-state matrices (Figs.~\ref{fig:clusteratlas}--\ref{fig:temporalsummary}) provide complementary descriptions. They use complete trajectories and should not be interpreted as evidence of early predictive performance. The actual held-out evaluation remains the nested classifier: Table~\ref{tab:classifier_summary} reports its observed range and baseline-relative differences. This separation prevents visually appealing full-trajectory structure from being substituted for a leakage-controlled generalization result.

\section{Discussion and Limitations}
\subsection{What the Measurements Support}
The benchmark separates ordinary task learning, partial fitting of a stable arbitrary assignment, and training under repeatedly reassigned targets. Their optimization outcomes and some topological summaries differ descriptively. A possible explanation is that informative targets support different directional organization than arbitrary assignments. The current measurements do not distinguish this explanation from rank, distance, scale, inactive-unit, or other geometric effects. We did not perform a matched-geometry null experiment or a causal intervention on topology.

Raw-versus-normalized disagreement is informative because it exposes dependence on representation. Normalization removes row magnitude, but can neither establish a functional interpretation nor make all layers comparable. ReLU symmetries can preserve outputs while changing coordinated weight geometry~\cite{dinh2017}. An incoming-row normalization alone does not remove effects of all compensating scalings in adjacent layers. Activation or function-based representations would be useful complementary measurements, but were not run.

The baseline result is equally important: a richly parameterized topological summary and visually distinguishable full-run trajectories do not imply incremental prediction. PCA and clustering use all observed epochs descriptively; the early classifier uses only epochs 1--5. They answer different questions. A negative small-sample classifier result neither establishes the universal uselessness of topology nor justifies claiming that topology is predictive.

\subsection{Limits of Scope and Inference}
Five seeds give coarse exact-test resolution and uncertain interval estimates. All configured tests fail conventional significance; this is not an equivalence finding. Unadjusted $t$ intervals that exclude zero reflect a different assumption and multiplicity treatment from the exact/Holm analysis. Inference cannot be strengthened by counting epochs, diagram intervals, or plotted points as independent seeds.

Only one MLP, one dataset, and AdamW are studied. The arbitrary-label condition does not fully memorize within the equal 100-epoch budget. Extending only that condition would change the comparison; no such extension is claimed. The changing assignment is deliberately extreme and does not establish a general result about gradual drift, realistic relabeling, continual learning, or adaptation capacity.

Finite clouds, fixed neuron sampling, unequal layer widths, and the selected filtration conventions constrain interpretation. The H1 feature cap induces concentrated early missingness, and bottleneck analysis remains absent. There are no activation-topology, architecture, optimizer, or sampling-size replications in the empirical results. Later multidimensional visualizations are exploratory reanalyses of the same 15 models, not independent confirmation. Future work should first test representation invariance, strengthen geometry-matched baselines, increase independent seeds, and evaluate controlled adaptation tasks before expanding scientific claims.

\section{Conclusion}
This study characterizes persistent homology of incoming-weight point clouds under correct, fixed-random, and changing-random supervision using a controlled paired design. All 15 models completed training and all 3,300 topology cases are available. Correct-label models generalize, fixed-random models achieve partial arbitrary-target fitting, and changing-random models remain at chance under the tested setup. Normalized fc3 H1 persistence differs descriptively, but configured exact/Holm comparisons do not establish conventional significance. Early topological features do not improve condition prediction over the simpler baseline. The contribution is a reproducible empirical comparison and a transparent assessment of its limitations, leaving incremental predictive and mechanistic value unresolved.

\section*{Availability and AI Assistance}
Fashion-MNIST is publicly available~\cite{xiao2017}. The manuscript package includes editable sources, figure-generation code, and numerical tables derived from the archived experiment; the complete original result archive is retained separately because of its size. No public repository identifier was supplied. Generative AI assistance was used for software development, visualization, and editorial revision. Numerical findings are taken from archived measurements rather than generated experimental observations. This study uses a public clothing-image benchmark and does not involve a participant study.

\clearpage
\appendices
\section{Reproducibility and Numerical Environment}
\label{app:repro}
The execution manifest records Linux, Python 3.13.16, a single NVIDIA Tesla T4, PyTorch 2.11.0+cu130, CUDA 13.0, torchvision 0.26.0+cu130, NumPy 2.4.6, SciPy 1.16.3, pandas 2.2.3, Ripser.py 0.6.15, Persim 0.4.1, scikit-learn 1.6.1, and Matplotlib 3.10.0. These are the archived \emph{experiment} versions; later manuscript rendering uses a separate recorded environment. Training uses one CUDA device; Ripser and matching use CPU. GPU memory is PyTorch peak allocated memory, not total device occupancy.

The configuration fixes the model seeds and partition seed. Initialization uses a seeded PyTorch generator within an isolated RNG context. NumPy SeedSequence keys $(s,29,e)$ specify minibatch order; $(s,17,e)$ specify moving-label permutations, and $(s,17,0)$ specifies the fixed permutation. Model loading occurs before constructing independent optimizer evolution. The setup enables deterministic PyTorch algorithms, disables cuDNN benchmarking, and sets a deterministic cuBLAS workspace configuration. Bitwise portability across different CUDA/PyTorch versions is not asserted.

The archive preserves configuration, environment, protocol, histories, target hashes, diagrams, cloud arrays, summaries, predictions, statistical results, initial states, selected epoch weights, and rolling resumable states. The compact export does not contain every per-epoch weight file, although the full runtime pipeline saves all of them. The data tensors/cache are also excluded from the compact export. Reproducing preprocessing requires regenerating the documented shared split from the official data. A separate machine-readable audit accompanies this paper.

\section{Data Partition and Supervision Audit}
\label{app:data}
Stratification with partition seed 314159 yields 600 validation and 5,400 fitting examples per class from the balanced official training set. Division by 255 precedes fitting-only mean/population-standard-deviation normalization. No augmentation or validation-dependent learning-rate policy is used. The original test labels remain unchanged in every condition. Validation/test examples never receive shuffled fitting labels.

An assignment hash records the full fitting-label vector for each epoch. All five correct-label runs use their shared original vector. Each fixed-random run has one constant hash, while every changing-random run has 100 distinct hashes. Hashes support assignment reproducibility, but do not by themselves validate every possible property of the training implementation; the source code defines the separate streams and the permutation-over-examples operation.

Independent fitting targets coexist with original-label monitoring. In particular, the assigned-target accuracy of a fixed-random model quantifies arbitrary-target fitting, whereas its original-label accuracy quantifies alignment with the actual clothing task. Near-chance true-label performance does not imply that the assigned arbitrary targets were not partly fitted. Table~\ref{tab:perseed} retains all three accuracy roles.

\section{Training Verification and Complete Final Metrics}
\label{app:training}
A verification pass checks that every seed/condition has exactly epochs 1 through 100, no duplicated history epochs, finite monitored loss/norm values, and a final test record. The archived topology clouds at epoch 0 are identical across conditions within each seed/layer/representation, and the sampled indices are identical across all cases of a layer. These are direct artifact checks. Optimizer independence follows from the construction of a separate AdamW object in each run; we do not infer it merely from matching initialization clouds.

Checkpoint writes use a temporary file followed by replacement. Each epoch retains portable CPU model tensors and a rolling optimizer/model/history state; historical metadata is also recorded. A resumed epoch sequence regenerates labels and minibatches from seed and epoch. A server termination can still interrupt a write or an epoch; successfully saved Drive artifacts and a stable experiment configuration are prerequisites for recovery. The study does not claim an uninterrupted session or an independent replay of every GPU run.

\begin{table*}[!t]
\centering
\caption{Complete final performance by seed and supervision condition.}
\label{tab:perseed}
\footnotesize
\setlength{\tabcolsep}{4pt}
\begin{tabular}{rlrrrrr}
\toprule
Seed & Condition & Online loss & Assigned train (\%) & True train (\%) & True val. (\%) & True test (\%) \\
\midrule
42 & Correct & 0.0214 & 99.50 & 99.50 & 89.85 & 89.60 \\
42 & Fixed random & 1.0002 & 66.91 & 10.20 & 10.23 & 10.86 \\
42 & Changing random & 2.3027 & 10.00 & 10.00 & 10.00 & 10.00 \\
123 & Correct & 0.0210 & 99.29 & 99.29 & 89.52 & 89.23 \\
123 & Fixed random & 0.8170 & 74.19 & 10.32 & 10.52 & 10.95 \\
123 & Changing random & 2.3027 & 10.00 & 10.00 & 10.00 & 10.00 \\
456 & Correct & 0.0183 & 99.75 & 99.75 & 89.75 & 89.35 \\
456 & Fixed random & 0.9242 & 70.05 & 9.19 & 9.45 & 9.48 \\
456 & Changing random & 2.3027 & 10.00 & 10.00 & 10.00 & 10.00 \\
789 & Correct & 0.0165 & 99.31 & 99.31 & 89.77 & 89.10 \\
789 & Fixed random & 0.9780 & 67.75 & 9.84 & 10.03 & 10.02 \\
789 & Changing random & 2.3027 & 10.00 & 10.00 & 10.00 & 10.00 \\
2026 & Correct & 0.0180 & 99.67 & 99.67 & 90.03 & 89.65 \\
2026 & Fixed random & 1.1429 & 60.80 & 10.08 & 10.27 & 10.14 \\
2026 & Changing random & 2.3027 & 10.00 & 10.00 & 10.00 & 10.00 \\
\bottomrule
\end{tabular}
\end{table*}

\section{Topological Definitions and Filtration Conventions}
\label{app:definitions}
\subsection{Clouds and Normalization}
The raw point cloud contains selected rows of a hidden-layer matrix, with ambient dimensions 784, 1024, 512, 256, and 128 for fc1--fc5, respectively. Bias terms are not appended. The corresponding point counts are 128, 128, 128, 128, and 64. A zero incoming vector would remain zero under the $10^{-12}$ denominator guard; it would not become a unit vector. The saved-cloud audit finds no such sampled row, and the minimum raw sampled row norm is approximately 0.4293.

A row permutation of a full cloud does not change its Rips diagram. A fixed-index subsample, however, must be transported under a neuron permutation to preserve the sampled object. Similarly, row normalization does not eliminate every equivalent-model scaling because compensating transformations can alter coordinates in adjacent layers. No symmetry-invariant functional measurement is claimed.

\subsection{Finite Intervals and Betti Curves}
Each full finite-cloud Rips filtration has one essential H0 interval. It is excluded from Eq.~\eqref{eq:summary}, mean/maximum lifetime and matching, but remains alive in the H0 Betti curve. H1 intervals are finite when the complete finite-cloud filtration is computed. Positive finite lifetimes define feature count; there is no unreported lifetime significance threshold. Empty summaries take zero values. Entropy uses natural logarithms. Mean lifetime averages all finite intervals returned by Ripser; positive-lifetime count is reported separately.

For comparison across matched cases, each layer/representation has a shared grid. Normalized clouds have diameter at most 2, giving a fixed $[0,2]$ range. Raw grids use an endpoint 1.001 times the largest measured pairwise distance across all cases of that layer. Betti curves use 128 equally spaced values; landscapes use 32 values and ranks 1--3. The tent for interval $(b_j,d_j)$ is
\begin{equation}
 f_j(u)=\max\{0,\min(u-b_j,d_j-u)\},
\end{equation}
and $\lambda_m(u)$ is its $m$th largest value across intervals, padded with zero where necessary. The early classifier retains normalized landscapes and normalized Betti area, but excludes the \emph{raw} counterparts. It does not exclude every grid-based feature.

\subsection{Distance Units and Computational Scope}
Equation~\eqref{eq:wasserstein} specifies $p=1$ with ground norm $q=2$. Essential intervals are removed before matching. Diagram cardinality and cloud geometry affect matching sums, so neither dividing by elapsed epochs nor using normalized input rows makes all layers' sums interchangeable. The feature cap is applied to finite diagrams after essential-interval removal. Here H0 has at most 127 finite intervals and is never capped.

For $S=5$ seeds, $C=3$ conditions, $K=22$ checkpoints, $L=5$ layers, $R=2$ representations, and two homology dimensions, the evaluation count is
\begin{equation}
 2LR\left[SC(K-1)+SK\binom C2+CK\binom S2\right]=26100.
\end{equation}
This accounts for 6,300 temporal, 6,600 condition, and 13,200 seed comparisons. Of 13,050 H1 comparisons, 282 are unavailable, leaving 12,768 finite H1 values; all 13,050 H0 values are finite. Bottleneck distance would characterize a maximum matching displacement under its chosen ground norm, while Wasserstein accumulates displacements. One must not claim a numerically larger L-infinity bottleneck cost than an L2-ground matching sum merely by describing a single large feature movement. No bottleneck conclusion is drawn here.

\section{Complete Topological Summaries and Associations}
\label{app:topology}
Table~\ref{tab:topofinal} gives endpoint H0/H1 totals for every layer, condition, and representation. These final values are not substitutes for Eq.~\eqref{eq:average}. Full per-case summaries remain in \texttt{topology.csv}; the paper package includes that machine-readable table rather than printing 3,300 rows and hundreds of landscape columns.

\begin{table*}[!t]
\centering
\caption{Layer-resolved H0 and H1 total persistence at epoch 100.}
\label{tab:topofinal}
\footnotesize
\setlength{\tabcolsep}{4pt}
\begin{tabular}{llrrrr}
\toprule
Layer & Condition & Raw H0 & Raw H1 & Normalized H0 & Normalized H1 \\
\midrule
fc1 & Correct & $343.099 \pm 12.210$ & $0.141 \pm 0.178$ & $131.867 \pm 1.510$ & $2.228 \pm 0.225$ \\
fc1 & Fixed random & $338.433 \pm 20.503$ & $0.101 \pm 0.092$ & $125.341 \pm 3.856$ & $0.528 \pm 0.152$ \\
fc1 & Changing random & $101.190 \pm 6.470$ & $0.299 \pm 0.093$ & $138.228 \pm 6.754$ & $0.316 \pm 0.180$ \\
fc2 & Correct & $394.241 \pm 6.829$ & $0.027 \pm 0.061$ & $149.362 \pm 1.108$ & $3.929 \pm 0.233$ \\
fc2 & Fixed random & $329.587 \pm 31.240$ & $0.015 \pm 0.020$ & $156.419 \pm 0.241$ & $1.738 \pm 0.189$ \\
fc2 & Changing random & $104.294 \pm 4.196$ & $0.021 \pm 0.019$ & $158.539 \pm 1.234$ & $1.953 \pm 0.244$ \\
fc3 & Correct & $321.424 \pm 5.234$ & $0.876 \pm 0.338$ & $134.333 \pm 0.898$ & $4.376 \pm 0.359$ \\
fc3 & Fixed random & $381.624 \pm 18.698$ & $0.003 \pm 0.007$ & $155.760 \pm 1.408$ & $2.755 \pm 0.395$ \\
fc3 & Changing random & $101.100 \pm 3.442$ & $0.038 \pm 0.033$ & $153.862 \pm 1.724$ & $2.352 \pm 0.461$ \\
fc4 & Correct & $196.466 \pm 1.808$ & $1.244 \pm 0.532$ & $125.874 \pm 1.619$ & $4.619 \pm 0.396$ \\
fc4 & Fixed random & $342.598 \pm 26.151$ & $0.021 \pm 0.031$ & $150.343 \pm 0.934$ & $6.034 \pm 0.838$ \\
fc4 & Changing random & $89.549 \pm 3.111$ & $0.271 \pm 0.223$ & $152.187 \pm 2.182$ & $3.836 \pm 0.472$ \\
fc5 & Correct & $64.447 \pm 2.256$ & $0.357 \pm 0.305$ & $68.621 \pm 0.628$ & $2.889 \pm 0.257$ \\
fc5 & Fixed random & $132.149 \pm 3.362$ & $0.098 \pm 0.030$ & $70.966 \pm 0.613$ & $3.303 \pm 0.400$ \\
fc5 & Changing random & $38.350 \pm 0.447$ & $0.493 \pm 0.063$ & $74.585 \pm 1.278$ & $2.326 \pm 0.424$ \\
\bottomrule
\end{tabular}
\par\vspace{2pt}\begin{minipage}{\textwidth}\scriptsize Means $\pm$ sample SD across five seeds. These are exploratory final-checkpoint summaries, not the time-averaged primary endpoint. Each representation has its own filtration units. fc1--fc4 contain 128 sampled neurons; fc5 contains 64.\end{minipage}
\end{table*}

\begin{table*}[!t]
\centering
\caption{Exploratory normalized fc3 H1 change-rate associations.}
\label{tab:corr}
\footnotesize
\setlength{\tabcolsep}{4pt}
\begin{tabular}{llrrrrr}
\toprule
Condition & Lag & Loss & Gradient & Update & Val. accuracy & Weight norm \\
\midrule
Correct & 0 & -0.096 & -0.099 & +0.242 & -0.063 & -0.042 \\
Correct & 1 & +0.040 & +0.158 & +0.219 & -0.212 & +0.026 \\
Fixed random & 0 & -0.358 & +0.418 & -0.026 & +0.032 & +0.138 \\
Fixed random & 1 & -0.393 & +0.505 & +0.028 & +0.172 & +0.044 \\
Changing random & 0 & +0.325 & +0.209 & +0.167 & -0.055 & -0.570 \\
Changing random & 1 & +0.112 & +0.391 & +0.089 & +0.092 & -0.498 \\
\bottomrule
\end{tabular}
\par\vspace{2pt}\begin{minipage}{\textwidth}\scriptsize Median within-seed Spearman correlation across five seeds. Lag 1 shifts the topological change-rate sequence by one sampled checkpoint; it is not a fixed one-epoch delay. No p-values or multiplicity-adjusted causal claims are inferred.\end{minipage}
\end{table*}

The archived exploratory association table contains 4,500 rows: five seeds, three conditions, five layers, two representations, three topological summaries, five monitored metrics, and two lags. It uses Spearman correlation of checkpoint rates after joining with monitored training epochs. Initialization has no training metric row and is not part of that join. Lag 0 aligns rate sequences; lag 1 shifts the topology-rate sequence once. No additional time-window restriction or higher-order detrending was applied. Undefined or insufficient correlations would be omitted rather than set to zero. Table~\ref{tab:corr} reports medians over the five seed-level H1-total correlations, not a correlation obtained by pooling every epoch from every model.

\subsection{Training and Optimization Milestones}
Tables~\ref{tab:milestones} and~\ref{tab:optimization} report the same five paired runs at epochs 1, 5, 10, 50, and 100. Reporting both assigned-target accuracy and true-label validation accuracy separates fitting randomized supervision from learning the original task. The true-label training--validation gap is computed within seed before aggregation. In particular, high assigned-target accuracy in the fixed-random condition is not evidence of true-label generalization. The online loss averages training batches under evolving parameters, whereas the reported accuracy evaluates the end-of-epoch model; these quantities need not satisfy a deterministic one-to-one relation.

\begin{table*}[!t]
\centering
\caption{Training milestones across all five seeds.}
\label{tab:milestones}
\footnotesize
\begin{tabular}{@{}llllll@{}}
\toprule
Epoch & Supervision & Online loss & Target acc. (\%) & Validation (\%) & True-label gap (pp) \\
\midrule
1 & Correct & $0.639 \pm 0.018$ & $84.064 \pm 0.776$ & $83.810 \pm 1.075$ & $0.254 \pm 0.349$ \\
1 & Fixed random & $2.303 \pm 0.000$ & $10.087 \pm 0.083$ & $7.687 \pm 3.722$ & $-0.022 \pm 0.214$ \\
1 & Changing random & $2.303 \pm 0.000$ & $10.090 \pm 0.120$ & $10.130 \pm 1.981$ & $-0.291 \pm 0.221$ \\
5 & Correct & $0.282 \pm 0.003$ & $90.453 \pm 0.301$ & $88.537 \pm 0.290$ & $1.916 \pm 0.226$ \\
5 & Fixed random & $2.303 \pm 0.000$ & $10.151 \pm 0.184$ & $11.623 \pm 3.157$ & $-0.066 \pm 0.280$ \\
5 & Changing random & $2.303 \pm 0.000$ & $10.175 \pm 0.073$ & $14.297 \pm 1.546$ & $-0.052 \pm 0.447$ \\
10 & Correct & $0.206 \pm 0.002$ & $92.931 \pm 0.326$ & $89.193 \pm 0.231$ & $3.738 \pm 0.196$ \\
10 & Fixed random & $2.303 \pm 0.000$ & $10.205 \pm 0.303$ & $9.413 \pm 5.484$ & $0.016 \pm 0.040$ \\
10 & Changing random & $2.303 \pm 0.000$ & $10.064 \pm 0.136$ & $10.460 \pm 2.821$ & $0.074 \pm 0.207$ \\
50 & Correct & $0.040 \pm 0.003$ & $98.766 \pm 0.098$ & $89.830 \pm 0.274$ & $8.936 \pm 0.177$ \\
50 & Fixed random & $1.909 \pm 0.170$ & $30.127 \pm 8.184$ & $10.690 \pm 1.427$ & $-0.131 \pm 0.609$ \\
50 & Changing random & $2.303 \pm 0.000$ & $10.000 \pm 0.000$ & $10.000 \pm 0.000$ & $0.000 \pm 0.000$ \\
100 & Correct & $0.019 \pm 0.002$ & $99.506 \pm 0.206$ & $89.783 \pm 0.187$ & $9.722 \pm 0.175$ \\
100 & Fixed random & $0.972 \pm 0.119$ & $67.938 \pm 4.887$ & $10.100 \pm 0.402$ & $-0.174 \pm 0.085$ \\
100 & Changing random & $2.303 \pm 0.000$ & $10.000 \pm 0.000$ & $10.000 \pm 0.000$ & $0.000 \pm 0.000$ \\
\bottomrule
\end{tabular}
\par\smallskip\begin{minipage}{0.97\textwidth}\footnotesize Entries are mean $\pm$ sample SD across five seeds. The gap is true-label training accuracy minus validation accuracy within each seed; it differs from assigned-target accuracy under randomized labels.\end{minipage}
\end{table*}

\begin{table*}[!t]
\centering
\caption{Optimization milestones and recorded epoch duration.}
\label{tab:optimization}
\footnotesize
\begin{tabular}{@{}llllll@{}}
\toprule
Epoch & Supervision & Gradient norm & Weight norm & Update norm & Epoch duration (s) \\
\midrule
1 & Correct & $4.299 \pm 0.697$ & $32.285 \pm 0.120$ & $18.533 \pm 0.217$ & $1.747 \pm 0.338$ \\
1 & Fixed random & $0.080 \pm 0.003$ & $27.177 \pm 0.067$ & $9.211 \pm 0.167$ & $1.826 \pm 0.344$ \\
1 & Changing random & $0.081 \pm 0.003$ & $27.223 \pm 0.106$ & $9.360 \pm 0.324$ & $1.864 \pm 0.385$ \\
5 & Correct & $1.001 \pm 0.047$ & $46.000 \pm 0.222$ & $9.249 \pm 0.235$ & $1.862 \pm 0.362$ \\
5 & Fixed random & $0.067 \pm 0.001$ & $30.603 \pm 1.010$ & $7.673 \pm 1.368$ & $1.598 \pm 0.060$ \\
5 & Changing random & $0.066 \pm 0.001$ & $29.563 \pm 0.254$ & $6.177 \pm 0.404$ & $1.659 \pm 0.193$ \\
10 & Correct & $0.827 \pm 0.043$ & $57.663 \pm 0.340$ & $9.491 \pm 0.114$ & $1.851 \pm 0.468$ \\
10 & Fixed random & $0.067 \pm 0.003$ & $36.059 \pm 2.391$ & $7.978 \pm 1.536$ & $2.034 \pm 0.364$ \\
10 & Changing random & $0.064 \pm 0.002$ & $32.420 \pm 1.051$ & $6.017 \pm 0.699$ & $1.844 \pm 0.422$ \\
50 & Correct & $0.518 \pm 0.013$ & $102.751 \pm 0.636$ & $10.353 \pm 1.391$ & $1.585 \pm 0.098$ \\
50 & Fixed random & $1.172 \pm 0.361$ & $108.203 \pm 5.471$ & $10.644 \pm 1.488$ & $1.911 \pm 0.510$ \\
50 & Changing random & $0.058 \pm 0.001$ & $36.574 \pm 2.436$ & $0.089 \pm 0.025$ & $1.567 \pm 0.080$ \\
100 & Correct & $0.358 \pm 0.021$ & $128.692 \pm 0.855$ & $10.389 \pm 2.213$ & $1.861 \pm 0.477$ \\
100 & Fixed random & $3.745 \pm 0.352$ & $135.932 \pm 3.543$ & $7.988 \pm 0.351$ & $1.762 \pm 0.338$ \\
100 & Changing random & $0.057 \pm 0.001$ & $34.236 \pm 3.008$ & $0.083 \pm 0.023$ & $1.738 \pm 0.214$ \\
\bottomrule
\end{tabular}
\par\smallskip\begin{minipage}{0.97\textwidth}\footnotesize Gradient norm is the monitored pre-clipping norm. Weight and update norms are global Euclidean norms. Recorded epoch duration excludes checkpoint I/O; no full-pipeline runtime is inferred.\end{minipage}
\end{table*}

\subsection{Layer and Representation Sensitivity}
Table~\ref{tab:layer_averages} extends the primary fc3 summary to every hidden layer. Trapezoidal weighting accounts for the dense early schedule and sparse late schedule, preventing the first ten epochs from receiving disproportionate weight simply because they were sampled more frequently. These additional layer comparisons were not added to the seven-outcome primary testing family; they are descriptive sensitivity analyses.

\begin{table*}[!t]
\centering
\caption{Time-weighted H1 total persistence by layer and representation.}
\label{tab:layer_averages}
\footnotesize
\begin{tabular}{@{}llll@{}}
\toprule
Layer & Supervision & Raw & Normalized \\
\midrule
fc1 & Correct & $0.281 \pm 0.087$ & $3.133 \pm 0.429$ \\
fc1 & Fixed random & $0.206 \pm 0.064$ & $0.848 \pm 0.126$ \\
fc1 & Changing random & $0.393 \pm 0.072$ & $0.364 \pm 0.102$ \\
fc2 & Correct & $0.302 \pm 0.086$ & $3.979 \pm 0.241$ \\
fc2 & Fixed random & $0.069 \pm 0.030$ & $1.608 \pm 0.146$ \\
fc2 & Changing random & $0.058 \pm 0.018$ & $1.973 \pm 0.172$ \\
fc3 & Correct & $1.314 \pm 0.175$ & $4.359 \pm 0.204$ \\
fc3 & Fixed random & $0.110 \pm 0.031$ & $2.743 \pm 0.227$ \\
fc3 & Changing random & $0.110 \pm 0.028$ & $2.575 \pm 0.408$ \\
fc4 & Correct & $1.554 \pm 0.318$ & $5.215 \pm 0.145$ \\
fc4 & Fixed random & $0.306 \pm 0.065$ & $5.134 \pm 0.540$ \\
fc4 & Changing random & $0.492 \pm 0.217$ & $4.115 \pm 0.378$ \\
fc5 & Correct & $0.686 \pm 0.084$ & $3.130 \pm 0.235$ \\
fc5 & Fixed random & $0.315 \pm 0.052$ & $2.724 \pm 0.219$ \\
fc5 & Changing random & $0.647 \pm 0.055$ & $2.422 \pm 0.396$ \\
\bottomrule
\end{tabular}
\par\smallskip\begin{minipage}{0.97\textwidth}\footnotesize Within each seed, trapezoidal integration over the actual 22 checkpoint epochs is divided by 100. Entries summarize these five seed-level values as mean $\pm$ sample SD. Cross-layer scales are not interchangeable.\end{minipage}
\end{table*}

\begin{table*}[!t]
\centering
\caption{Final fc3 geometry and finite H1 structure.}
\label{tab:final_geometry}
\footnotesize
\begin{tabular}{@{}lllllll@{}}
\toprule
Supervision & Cloud & Mean distance & Effective rank & Leading singular value & H1 count & Max. lifetime \\
\midrule
Correct & raw & $3.629 \pm 0.033$ & $107.556 \pm 0.580$ & $9.134 \pm 0.243$ & $12.000 \pm 3.162$ & $0.255 \pm 0.078$ \\
Correct & normalized & $1.352 \pm 0.003$ & $110.796 \pm 0.400$ & $3.577 \pm 0.079$ & $127.600 \pm 8.173$ & $0.169 \pm 0.018$ \\
Fixed random & raw & $4.287 \pm 0.191$ & $113.151 \pm 3.058$ & $8.393 \pm 0.438$ & $0.400 \pm 0.548$ & $0.003 \pm 0.007$ \\
Fixed random & normalized & $1.372 \pm 0.004$ & $119.148 \pm 0.528$ & $3.216 \pm 0.205$ & $169.600 \pm 19.282$ & $0.069 \pm 0.010$ \\
Changing random & raw & $0.939 \pm 0.047$ & $117.844 \pm 0.677$ & $3.008 \pm 0.316$ & $5.400 \pm 4.722$ & $0.012 \pm 0.008$ \\
Changing random & normalized & $1.319 \pm 0.015$ & $119.018 \pm 0.532$ & $4.314 \pm 0.273$ & $194.200 \pm 19.690$ & $0.050 \pm 0.009$ \\
\bottomrule
\end{tabular}
\par\smallskip\begin{minipage}{0.97\textwidth}\footnotesize All entries are mean $\pm$ sample SD across seeds. Spectral quantities use the sampled cloud, not the full weight matrix. Effective rank uses normalized singular values, not squared singular values.\end{minipage}
\end{table*}

\begin{table*}[!t]
\centering
\caption{Within-seed raw versus normalized H1 trajectory agreement.}
\label{tab:representation_agreement}
\footnotesize
\begin{tabular}{@{}lllll@{}}
\toprule
Layer & Supervision & Median Spearman & Minimum & Maximum \\
\midrule
fc1 & Correct & 0.630 & 0.421 & 0.759 \\
fc1 & Fixed random & 0.142 & -0.101 & 0.675 \\
fc1 & Changing random & 0.570 & 0.526 & 0.936 \\
fc2 & Correct & 0.173 & -0.185 & 0.311 \\
fc2 & Fixed random & 0.508 & 0.436 & 0.599 \\
fc2 & Changing random & 0.632 & -0.123 & 0.850 \\
fc3 & Correct & 0.469 & -0.136 & 0.630 \\
fc3 & Fixed random & 0.730 & 0.617 & 0.829 \\
fc3 & Changing random & 0.942 & 0.819 & 0.951 \\
fc4 & Correct & 0.623 & 0.225 & 0.845 \\
fc4 & Fixed random & 0.512 & 0.328 & 0.877 \\
fc4 & Changing random & 0.907 & 0.851 & 0.953 \\
fc5 & Correct & 0.571 & 0.362 & 0.851 \\
fc5 & Fixed random & 0.448 & 0.149 & 0.855 \\
fc5 & Changing random & 0.814 & 0.232 & 0.916 \\
\bottomrule
\end{tabular}
\par\smallskip\begin{minipage}{0.97\textwidth}\footnotesize Correlations use 22 matched checkpoints per seed. The range is across five seeds. These descriptive associations may reflect shared training trends; they are not independent-observation significance tests.\end{minipage}
\end{table*}

\begin{figure*}[!t]
\centering\includegraphics[width=\textwidth]{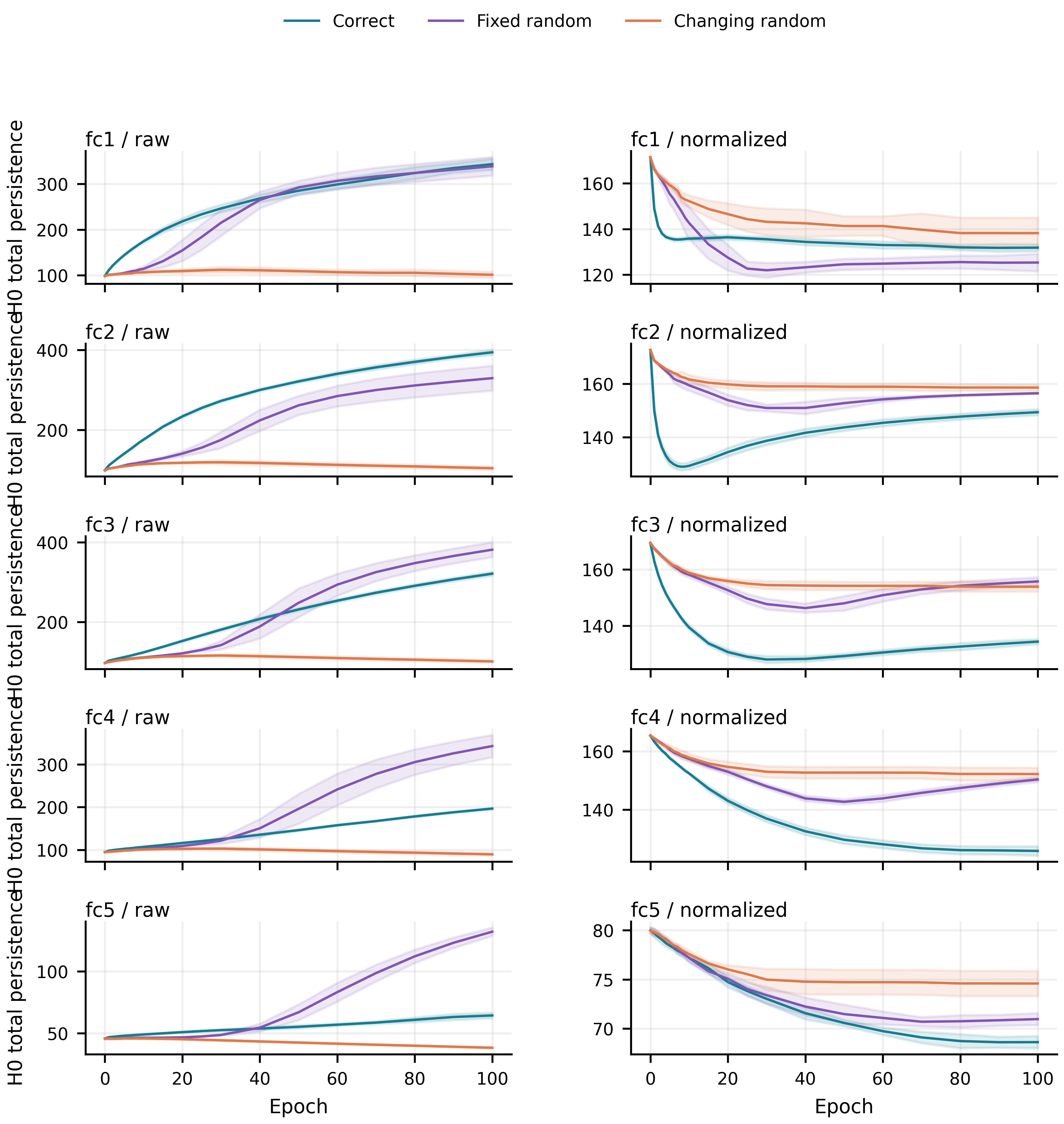}
\caption{H0 total persistence for all five hidden layers, in raw and normalized incoming-weight clouds. Lines are five-seed means; shading is one sample standard deviation. Colors follow the main-text supervision convention. Each panel retains its own vertical scale to expose within-layer evolution.}
\label{fig:h0all}
\end{figure*}

\begin{figure*}[!t]
\centering\includegraphics[width=\textwidth]{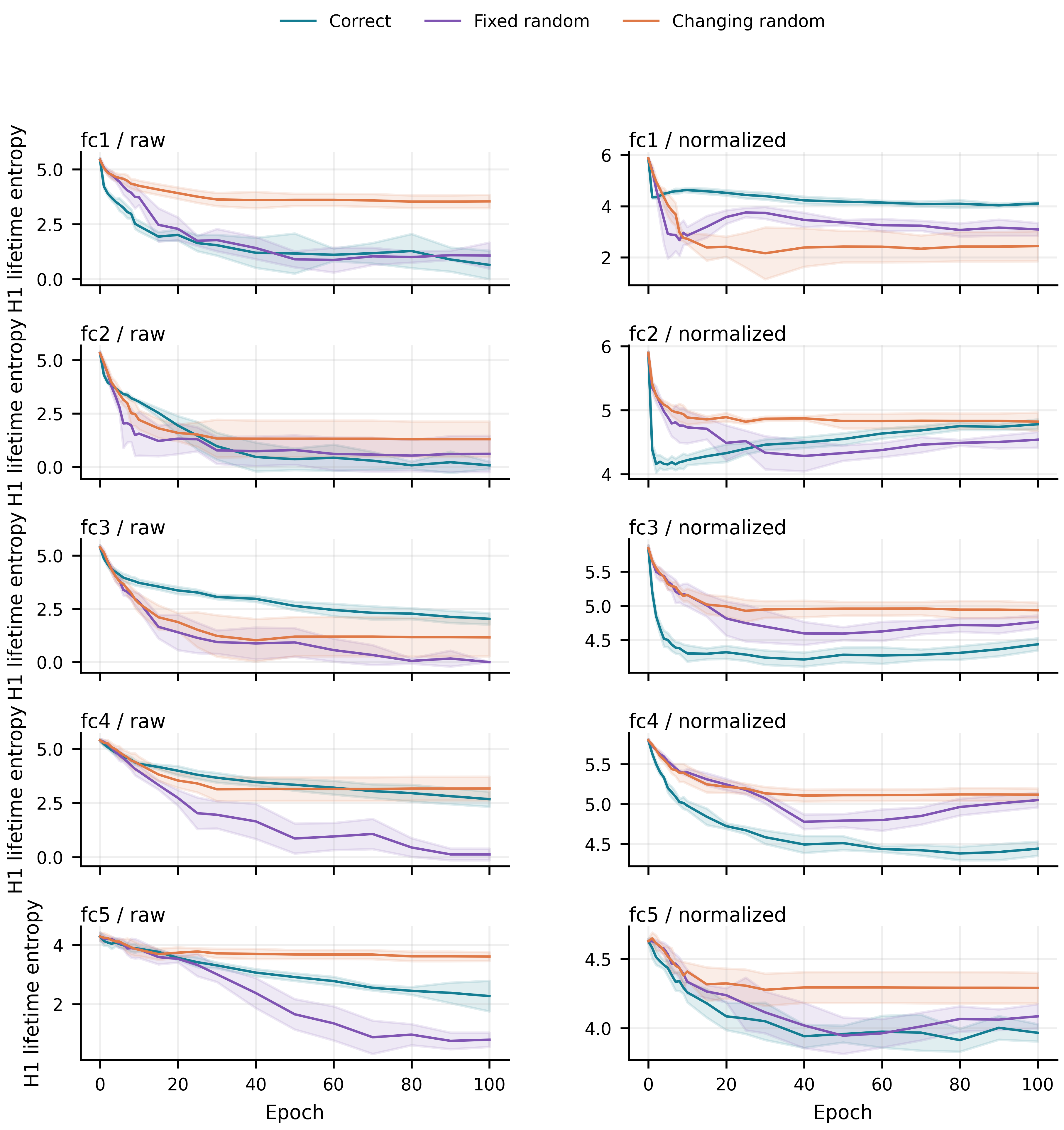}
\caption{Finite H1 lifetime entropy throughout training. Each row is one hidden layer; columns distinguish raw and normalized clouds. Five-seed mean and sample standard deviation are shown. Entropy is dimensionless and summarizes normalized lifetime weights, rather than the magnitude of persistence.}
\label{fig:entropyall}
\end{figure*}

Figures~\ref{fig:h0all} and~\ref{fig:entropyall} distinguish connectivity-scale changes from changes in the distribution of finite loop lifetimes. H0 total persistence aggregates finite merger scales, while lifetime entropy depends on relative lifetime weights. An entropy change therefore does not necessarily imply a corresponding change in total persistence. The shaded bands summarize between-seed variability at each checkpoint and are not confidence bands for independent epochs.

Table~\ref{tab:final_geometry} places endpoint topology beside sampled-cloud geometry. Row normalization controls individual row length but does not fix pairwise angular separation or the cloud's singular spectrum. Table~\ref{tab:representation_agreement} quantifies within-seed agreement of raw and normalized H1 trajectories. A large positive correlation can coexist with different absolute persistence levels; correlation is not an invariance guarantee and can arise from a common training trend.

\begin{figure*}[!t]
\centering\includegraphics[width=\textwidth]{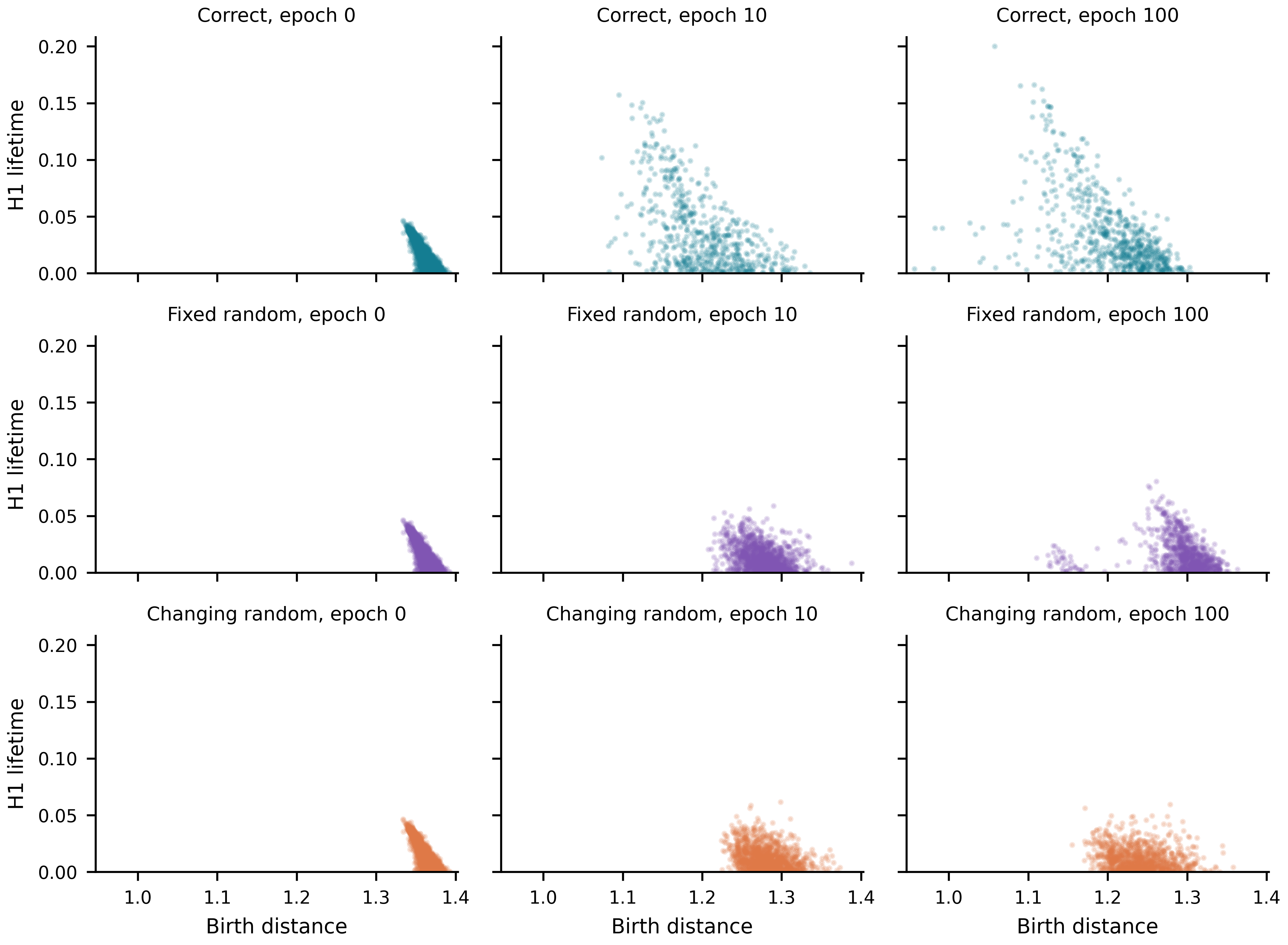}
\caption{Birth--lifetime views of all finite H1 intervals for normalized fc3 at epochs 0, 10, and 100. Each panel pools five seeds solely for visualization; points are not treated as independent experimental replicates. Axes are shared across the nine panels. Overplotting and diagram cardinality influence apparent density. Initialization panels coincide because paired clouds are identical within seed.}
\label{fig:diagramatlas}
\end{figure*}

Figure~\ref{fig:diagramatlas} shows how aggregate summaries relate to the actual intervals. Long-lived intervals appear farther above the horizontal axis, while horizontal displacement changes birth scale. A denser panel may contain more short intervals rather than stronger persistent structure. This view supplements lifetime totals without imposing a new significance threshold or deleting low-persistence features.

\section{Distance Coverage and Missingness}
\label{app:coverage}
Tables~\ref{tab:skips} and~\ref{tab:skipepoch} decompose every skipped matching calculation. All are normalized H1 comparisons in fc1--fc4, and all involve the first recorded epochs. Capping is therefore not missing completely at random with respect to diagram size or training time. A temporal endpoint averaged over available rates may differ from an endpoint defined on all intended intervals. This limitation is retained even though every point cloud and persistence diagram is present.

\begin{table}[!t]
\centering
\caption{Unavailable H1 distances by layer and comparison type.}
\label{tab:skips}
\footnotesize
\setlength{\tabcolsep}{4pt}
\begin{tabular}{lrrrr}
\toprule
Layer & Temporal & Conditions & Seeds & Total \\
\midrule
fc1 & 15 & 15 & 30 & 60 \\
fc2 & 15 & 15 & 30 & 60 \\
fc3 & 16 & 17 & 34 & 67 \\
fc4 & 21 & 26 & 48 & 95 \\
Total & 67 & 73 & 142 & 282 \\
\bottomrule
\end{tabular}
\par\vspace{2pt}\begin{minipage}{\columnwidth}\scriptsize All 282 skips are in normalized clouds; raw H1 and all H0 distances are available. No fc5 comparison was skipped. A skip occurs when either finite diagram contains more than 400 intervals.\end{minipage}
\end{table}

\begin{table}[!t]
\centering
\caption{Skipped H1 comparisons by recorded epoch.}
\label{tab:skipepoch}
\footnotesize
\setlength{\tabcolsep}{4pt}
\begin{tabular}{rrrrr}
\toprule
Epoch & Temporal & Conditions & Seeds & Total \\
\midrule
0 & 0 & 60 & 120 & 180 \\
1 & 60 & 13 & 22 & 95 \\
2 & 7 & 0 & 0 & 7 \\
\bottomrule
\end{tabular}
\par\vspace{2pt}\begin{minipage}{\columnwidth}\scriptsize Temporal rows are assigned to the later endpoint; condition and seed comparisons use their common endpoint. The decomposition does not count a missing distance as zero.\end{minipage}
\end{table}

Bottleneck values in the distance table are missing by design. A computationally deferred metric is not a measured zero. The numerical summaries, figures, and prediction inputs use the available Wasserstein values and identify caps in the status column. During classification, median imputation is learned within each training fold; it does not reclassify an unavailable raw distance as an experimentally observed zero.

\section{Complete Statistical Family and Per-Seed Effects}
\label{app:statistics}
The original configured family has seven tests, all listed in Table~\ref{tab:tests}. Three pairwise condition contrasts use time-averaged normalized fc3 H1 total persistence, three use the corresponding mean available temporal Wasserstein rate, and one compares topology and baseline classification. Table~\ref{tab:paired} gives the complete five-seed effect vectors, with orientations changed only for readability; reversing a contrast flips its mean and interval but not its two-sided $p$ value.

\begin{table*}[!t]
\centering
\caption{All per-seed effects in the configured comparison family.}
\label{tab:paired}
\footnotesize
\setlength{\tabcolsep}{4pt}
\begin{tabular}{llrrrrr}
\toprule
Outcome & Comparison & 42 & 123 & 456 & 789 & 2026 \\
\midrule
$\overline{\mathrm{TP}}_1$ & Fixed random $-$ Changing random & -0.0781 & 0.2515 & 0.2400 & 0.7685 & -0.3417 \\
$\overline{\mathrm{TP}}_1$ & Correct $-$ Fixed random & 1.3069 & 1.9862 & 1.5417 & 1.6765 & 1.5666 \\
$\overline{\mathrm{TP}}_1$ & Correct $-$ Changing random & 1.2287 & 2.2378 & 1.7817 & 2.4449 & 1.2249 \\
$\overline{r}_1$ & Fixed random $-$ Changing random & 0.2444 & 0.0586 & -0.0648 & 0.1408 & 0.0973 \\
$\overline{r}_1$ & Correct $-$ Fixed random & -0.0473 & -0.0007 & 0.0246 & -0.1077 & -0.0722 \\
$\overline{r}_1$ & Correct $-$ Changing random & 0.1970 & 0.0579 & -0.0402 & 0.0331 & 0.0251 \\
$\Delta\mathrm{BA}$ & Topology $-$ Baseline & 33.3333 & -33.3333 & 0.0000 & -33.3333 & 0.0000 \\
\bottomrule
\end{tabular}
\par\vspace{2pt}\begin{minipage}{\textwidth}\scriptsize Orientations match Table~\ref{tab:tests}; classification differences are percentage points. No epoch is used as an independent replicate.\end{minipage}
\end{table*}

For a test statistic $T=|\bar\delta|$, the exact sign-flip value is
\begin{equation}
 p=\frac1{32}\sum_{\epsilon\in\{-1,1\}^5}
 \mathbf 1\!\left\{\left|\frac15\sum_{s=1}^5\epsilon_s\delta_s\right|\geq|\bar\delta|\right\}.
\end{equation}
The implementation uses a $10^{-14}$ numerical tolerance in the comparison. Sign symmetry is an assumption of the null distribution, not something guaranteed merely by a paired design. Two-sided resolution is coarse. All five differences have the same sign in each correct-versus-random persistence contrast, producing $p=0.0625$. The two smallest values in the seven-test family become $0.4375$ after Holm monotonicity adjustment. Raw and adjusted values are computed from actual differences, not inferred from the adjusted result alone.

The intervals in Table~\ref{tab:tests} and Fig.~\ref{fig:effects} are $t_4$ intervals, not percentile-bootstrap or permutation confidence intervals. They are unadjusted and rely on a different distributional approximation. Figure~\ref{fig:effects} presents these intervals together with the individual seed effects to expose uncertainty rather than substituting parametric intervals for the configured exact tests. Outer-fold classification effects share fitted training observations across folds, so their reported interval and sign-flip value are descriptive.

\section{Prediction Features, Splits, and Regularization}
\label{app:prediction}
The nominal topology vector has 1,090 coordinates. Each of five normalized layers supplies ten finite-lifetime summaries (five each for H0 and H1), two Betti areas, and $2\times3\times32=192$ landscape coordinates, giving $5\times204=1020$. Five raw layers contribute only the ten finite-lifetime summaries, giving 50. Temporal Wasserstein-rate summaries for five layers, two representations, and two homology dimensions contribute 20. Each coordinate is averaged over the eligible early checkpoints rather than retaining five separate time positions.

The 94 baseline coordinates are nine geometric quantities for each of ten layer/representation combinations (90 coordinates) plus four training averages. Geometry comprises row-norm mean/SD; pairwise-distance mean/SD/maximum; three leading singular values; and effective rank of the sampled cloud matrix. If its singular values are $\sigma_i$, effective rank is
\begin{equation}
 \exp\!\left(-\sum_i q_i\log q_i\right),\qquad q_i=\sigma_i/\sum_j\sigma_j.
\end{equation}
This uses singular values, not squared energy fractions. The four training quantities are loss, pre-clipping gradient norm, global weight norm, and update norm. Neither validation/test accuracy nor class identity is a predictor. Combined dimensionality is 1,184. Constant or sparse coordinates are retained by the specified imputer/scaler rather than used to claim a lower-dimensional effective feature budget.

Each outer fold holds out one seed's three examples, leaving 12 training examples from four seeds. Inner leave-one-seed-out selection therefore fits on nine examples and validates on three at each split. The pipeline uses training-fold median imputation with empty-feature retention, StandardScaler, and multinomial logistic regression with L2 regularization, the L-BFGS solver, and 3,000 maximum iterations. The $C$ grid is $\{0.01,0.1,1,10\}$ and selection maximizes mean inner balanced accuracy. Table~\ref{tab:C} gives all selected values.

\begin{table}[!t]
\centering
\caption{Inner-validation choices of inverse regularization strength $C$.}
\label{tab:C}
\footnotesize
\setlength{\tabcolsep}{4pt}
\begin{tabular}{lrrr}
\toprule
Held-out seed & Topology & Baseline & Combined \\
\midrule
42 & 0.1 & 1 & 0.01 \\
123 & 0.01 & 0.1 & 0.01 \\
456 & 0.01 & 0.01 & 0.01 \\
789 & 0.01 & 0.01 & 0.01 \\
2026 & 0.01 & 0.01 & 0.01 \\
\bottomrule
\end{tabular}
\par\vspace{2pt}\begin{minipage}{\columnwidth}\scriptsize All searches used $C\in\{0.01,0.1,1,10\}$. Smaller $C$ means stronger L2 regularization; selection used only inner training folds.\end{minipage}
\end{table}

Topology has nine correct held-out condition assignments in aggregate, and baseline has ten, out of 15; combined also has nine. These totals follow from one example per class in each fold. The archive does not contain the individual predicted class labels, so no model-specific confusion matrix, classwise precision, or per-example explanation is fabricated. The saved fold table is the complete available predictive result. Detailed feature names and the reconstructed feature matrix are included in the source package; reconstruction does not refit classifiers or modify the recorded scores.

\begin{table*}[!t]
\centering
\caption{Descriptive summary of the five held-out-seed folds.}
\label{tab:classifier_summary}
\footnotesize
\begin{tabular}{@{}llllll@{}}
\toprule
Feature family & Coordinates & Balanced acc. (\%) & Minimum & Maximum & Versus baseline (pp) \\
\midrule
baseline & 94 & $66.667 \pm 23.570$ & 33.33 & 100.00 & $0.000 \pm 0.000$ \\
combined & 1184 & $60.000 \pm 14.907$ & 33.33 & 66.67 & $-6.667 \pm 27.889$ \\
topology & 1090 & $60.000 \pm 14.907$ & 33.33 & 66.67 & $-6.667 \pm 27.889$ \\
\bottomrule
\end{tabular}
\par\smallskip\begin{minipage}{0.97\textwidth}\footnotesize Mean $\pm$ sample SD describes the five folds; overlapping training sets prevent interpreting folds as independent replications. Each test fold contains three models. Per-class predictions were not saved, so no confusion matrix is reconstructed.\end{minipage}
\end{table*}

The descriptive range in Table~\ref{tab:classifier_summary} makes the coarseness of three-example test folds explicit: a single changed prediction moves a fold's balanced accuracy by 33.33 percentage points. The topology and combined feature families each average 60\%, versus 66.67\% for the geometric/optimization baseline. Neither the larger coordinate budget nor combining feature families produced an observed gain in this small evaluation. This does not establish that topology is uninformative in larger data regimes.

\section{Extended Visualizations and Their Interpretation}
\label{app:visual}
\begin{figure*}[!t]
\centering\includegraphics[width=\textwidth]{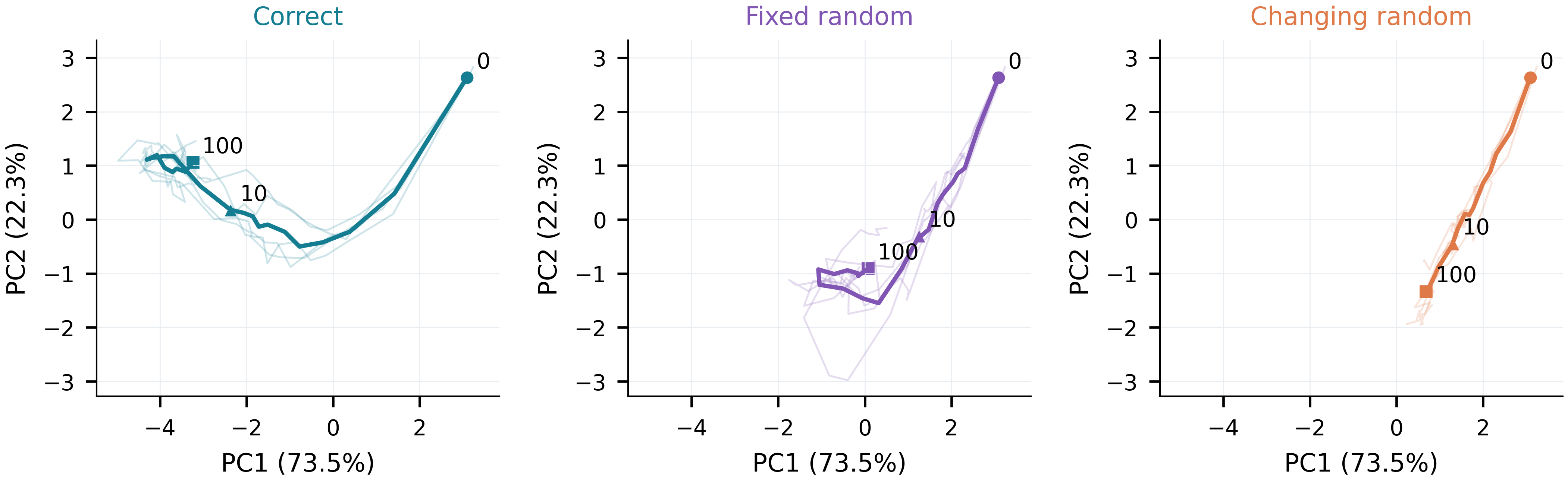}
\caption{Exploratory normalized fc3 topological-feature trajectories in a common PCA space. Six standardized summaries are used, with log transforms on magnitude/count coordinates. PC1 and PC2 explain 73.5\% and 22.3\% of the full descriptive variance, respectively. Thin lines show seeds, thick lines show condition means; circles, triangles, and squares identify epochs 0, 10, and 100. Apparent separation does not establish a mechanism, a phase transition, or held-out predictive gain.}
\label{fig:PCA}
\end{figure*}

Figure~\ref{fig:PCA} summarizes normalized fc3 trajectories using six features: H0 total persistence and entropy; H1 total persistence, entropy, maximum lifetime, and feature count. A $\log(1+x)$ transform is applied to the nonnegative magnitude/count coordinates, and all six features are standardized jointly over all saved seed/condition/checkpoint observations. Full-data PCA is a descriptive visualization, not part of held-out early prediction. Thin paths show seeds, thick paths show condition means, and endpoint markers identify measured epochs.

\begin{figure*}[!t]
\centering\includegraphics[width=\textwidth]{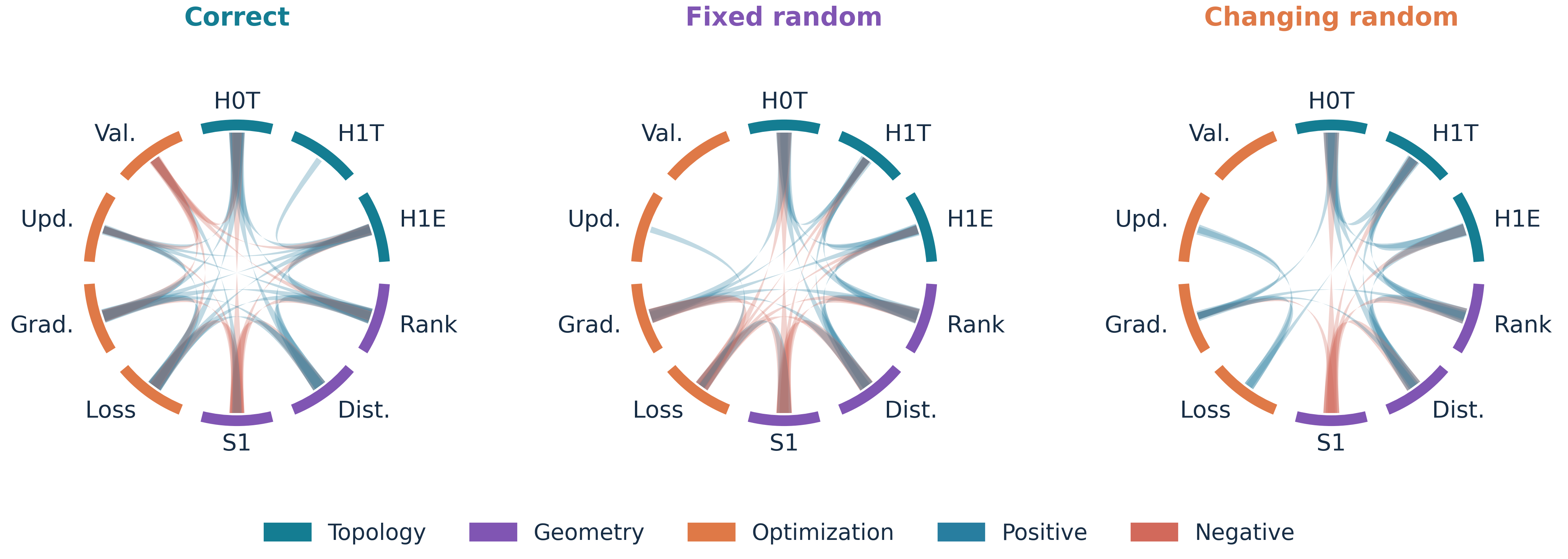}
\caption{Exploratory association ribbons for normalized fc3. H0T/H1T: total persistence; H1E: entropy; Rank: effective rank; Dist.: mean pairwise distance; S1: leading singular value; Loss: online loss; Grad.: pre-clipping gradient norm; Upd.: epoch update norm; Val.: validation accuracy. Ribbons require $|\mathrm{median}\,\rho|\geq0.25$ and sign agreement in at least four seeds. They represent undirected associations, not neuron connectivity or causal circuits.}
\label{fig:chords}
\end{figure*}

Figure~\ref{fig:chords} links topological, geometric, and optimization change-rate diagnostics. Each edge uses a median within-seed Spearman coefficient, an absolute median threshold of 0.25, and at least four same-sign seeds. Width represents coefficient magnitude; blue and red indicate positive and negative association. These are exploratory plotting thresholds, not significance thresholds or directed effects. Leading singular value is used instead of a normalized row-norm diagnostic that is nearly constant by construction.

\begin{figure*}[!t]
\centering\includegraphics[width=\textwidth]{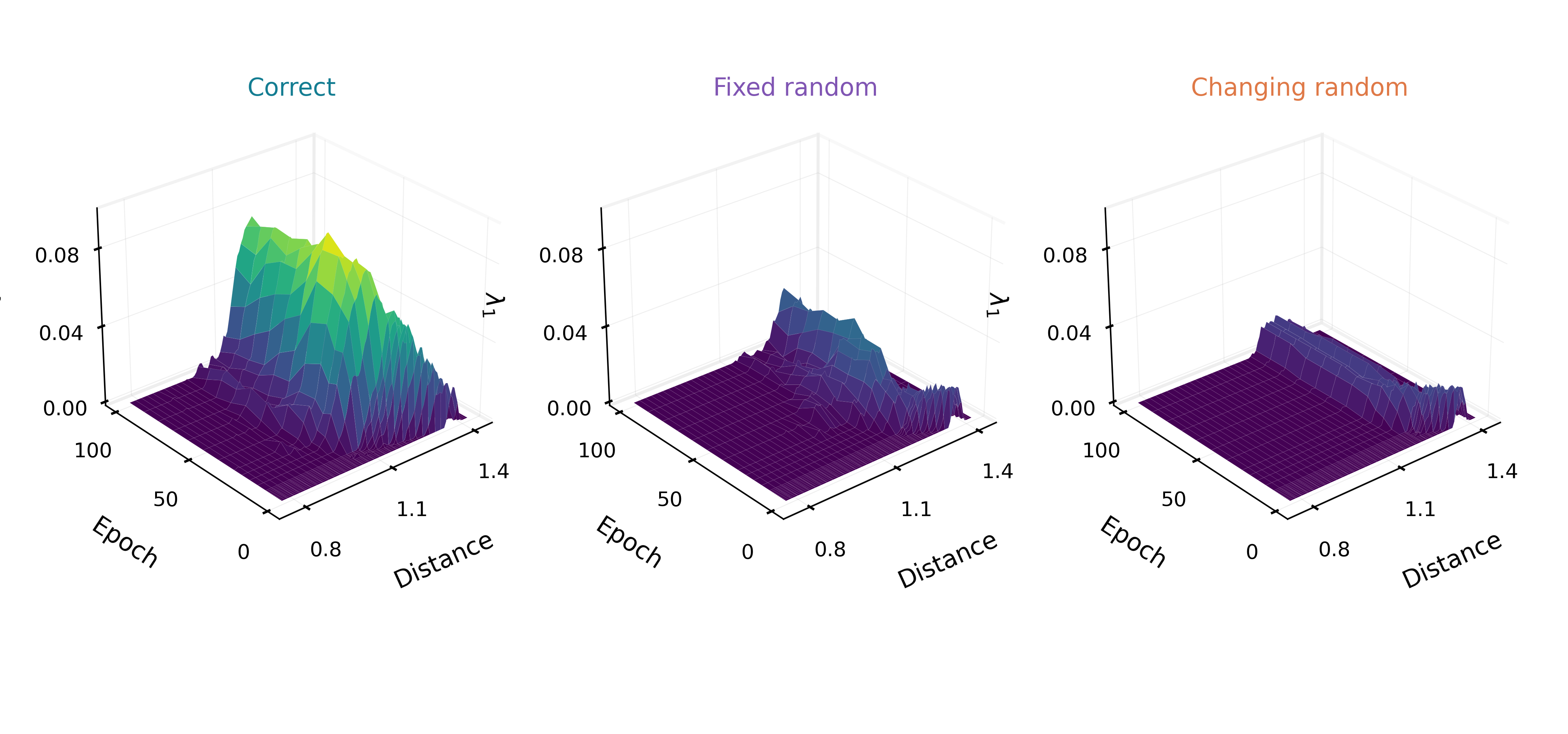}
\caption{Exploratory rank-one H1 persistence landscapes in normalized fc3 clouds. Heights average the first landscape separately computed for each seed. A common display grid and height scale are used; epoch positions use actual training time. The visualization includes all finite diagrams, even when a corresponding matching calculation exceeded the distance cap. Surfaces connecting sampled epochs are graphical interpolation only.}
\label{fig:landscapes}
\end{figure*}

Figure~\ref{fig:landscapes} provides rank-one persistence landscape surfaces on a shared display grid calculated from the saved finite H1 diagrams. Averaging is performed over seeds after computing each diagram's tent envelope. Surface polygons connect measured epochs for display and do not create new observations. Both axis units and height scale are shared across conditions.

The accompanying exploratory figure collection additionally contains temporal feature-distance matrices, layer-time contrasts, hierarchy-based signature plots, multivariate scatter matrices, three-dimensional dispersion ellipsoids, and circular atlases. Their conventions are documented in their captions. Standardized heatmaps do not make layer filtrations physically equivalent. Ellipsoids of five seed points describe covariance geometry, not a 95\% confidence region. Average-linkage clustering and full-data PCA do not supply independent replications. Visualization complexity does not add experimental evidence.

\subsection{Geometry--Topology Trajectories}
\begin{figure*}[!t]
\centering\includegraphics[width=\textwidth]{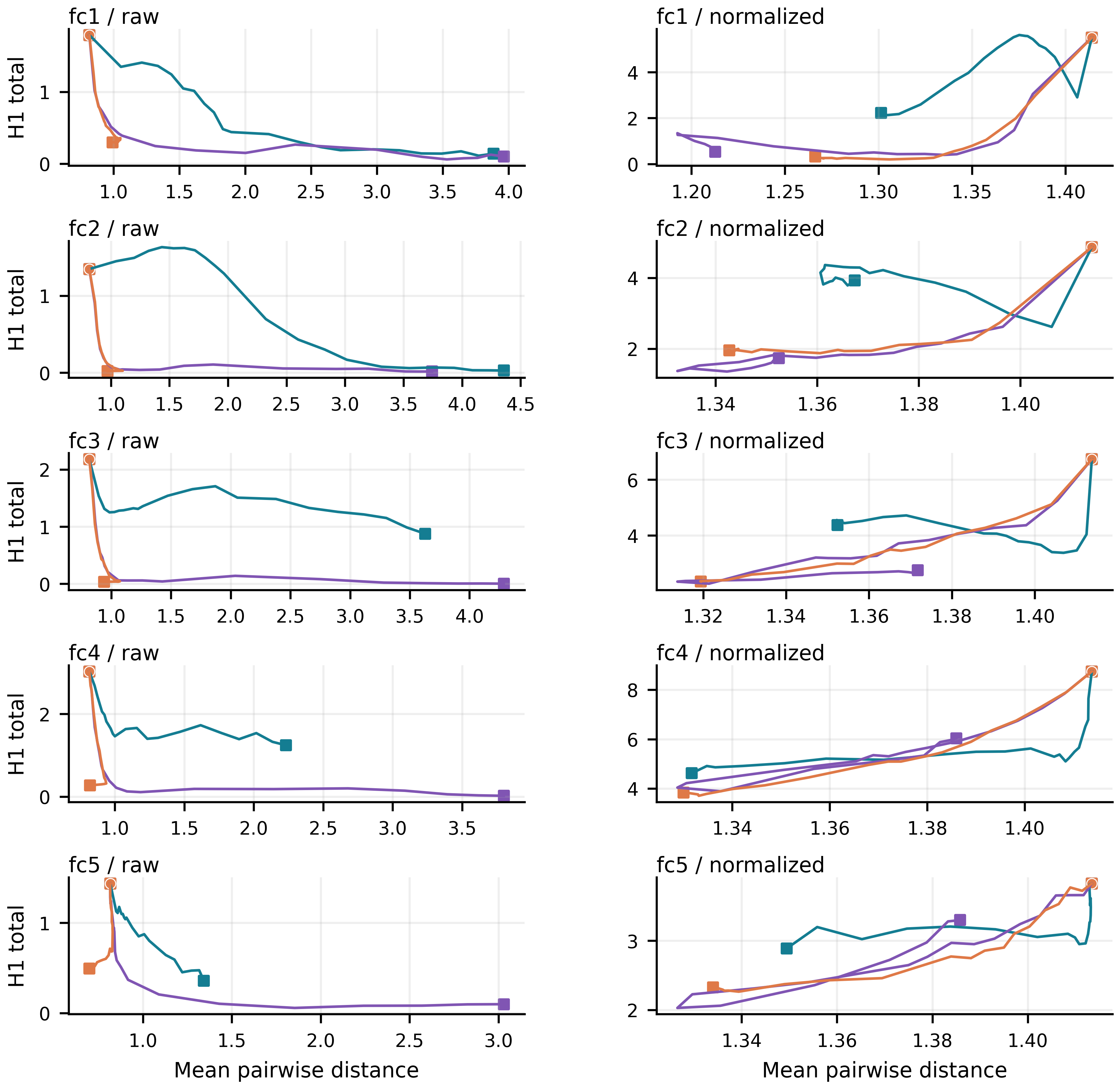}
\caption{Five-seed mean trajectories of H1 total persistence against mean pairwise cloud distance, for every hidden layer and both representations. Circles mark initialization and squares mark epoch 100; connected intermediate points follow the saved checkpoint schedule. Panel scales vary. Curves are descriptive geometric paths, not a fitted dynamical-system model.}
\label{fig:phaseatlas}
\end{figure*}
Figure~\ref{fig:phaseatlas} tests a visual interpretation of the topology curves against an elementary geometric scale. Normalization removes row magnitude, but a normalized trajectory may still move in mean pairwise distance. A bend or loop in this projection does not establish hysteresis, an attractor, or chaos: the visualization omits other state variables and averages distinct seed trajectories.

\subsection{Multivariate Seed-Level Signatures}
\begin{figure*}[!t]
\centering\includegraphics[width=\textwidth]{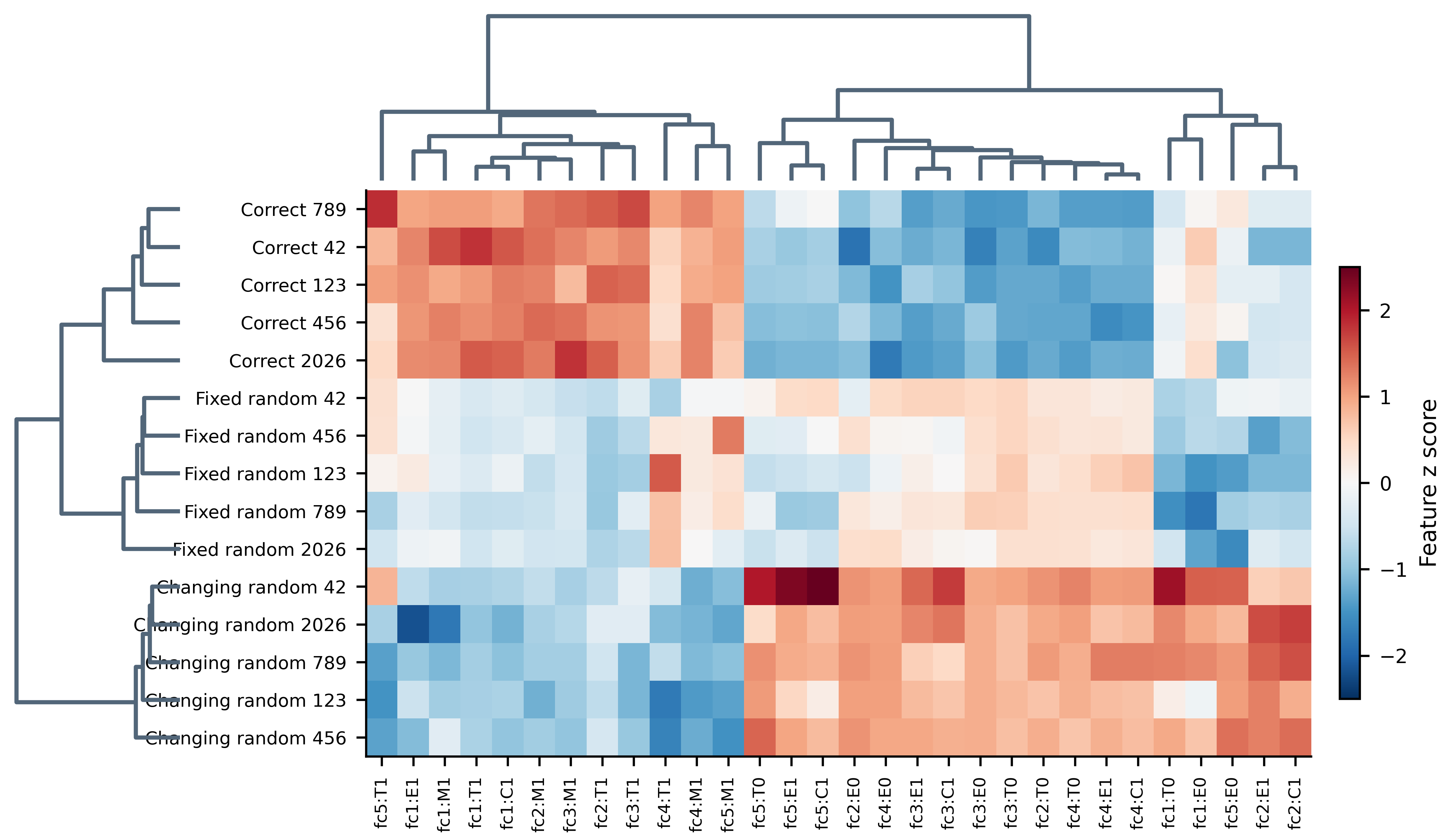}
\caption{Exploratory hierarchical clustering of 15 normalized-cloud run signatures. Each of 30 columns is a time-weighted summary (five layers times six metrics). T0/T1: H0/H1 total persistence; E0/E1: lifetime entropy; M1: maximum H1 lifetime; C1: finite H1 count. Columns are standardized across the 15 runs; average-linkage Euclidean clustering determines row and column order. Color saturates at $\pm2.5$ standard deviations.}
\label{fig:clusteratlas}
\end{figure*}
The clustering in Fig.~\ref{fig:clusteratlas} summarizes complete trajectories, including late training. It is not an early-prediction evaluation and its full-data standardization must not be reused as a preprocessing step in the held-out classifier. Feature-wise standardization emphasizes relative run variation and removes absolute cross-feature units; the dendrogram is not an uncertainty estimate or a phylogenetic relationship.

\subsection{Layer-Wise Association Structure}
\begin{figure*}[!t]
\centering\includegraphics[width=\textwidth]{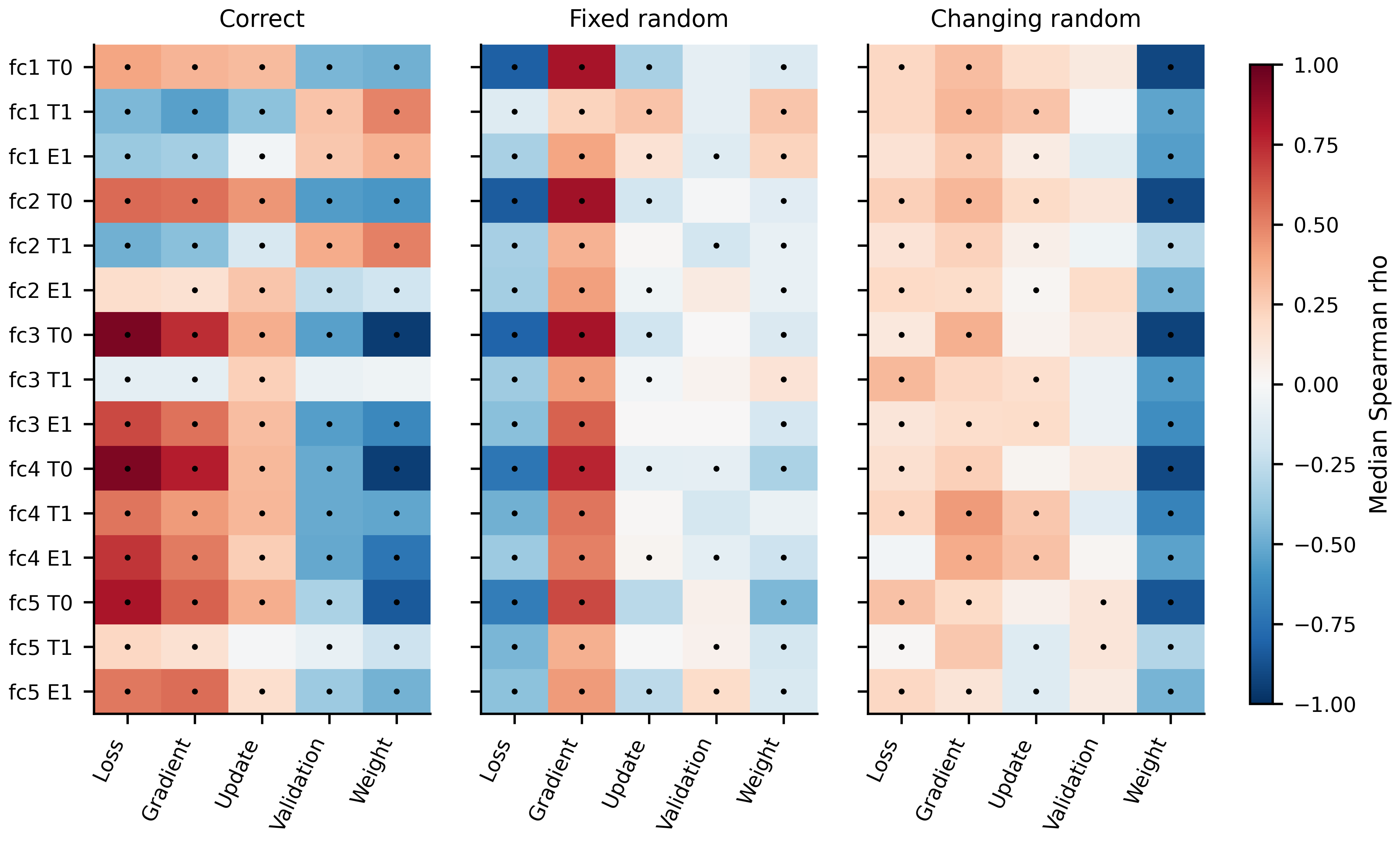}
\caption{Median seed-level Spearman associations between normalized-cloud topology change rates and monitored metric change rates at lag 0. Each cell summarizes five seeds. A small dot marks at least four same-sign seed correlations, without implying statistical significance. T0/T1 denote H0/H1 total persistence; E1 denotes H1 entropy. Shared color limits permit comparisons across supervision conditions.}
\label{fig:associationatlas}
\end{figure*}
Figure~\ref{fig:associationatlas} expands the association results across depth and topological summaries. The input is the archived change-rate association table, not a new pooled correlation over all model checkpoints. Sign agreement is a descriptive reproducibility cue. Temporal dependence, numerous displayed cells, and the shared influence of training time prevent interpreting isolated large correlations as corrected discoveries or causal links.

\subsection{Temporal Separation of Summary States}
\begin{figure*}[!t]
\centering\includegraphics[width=\textwidth]{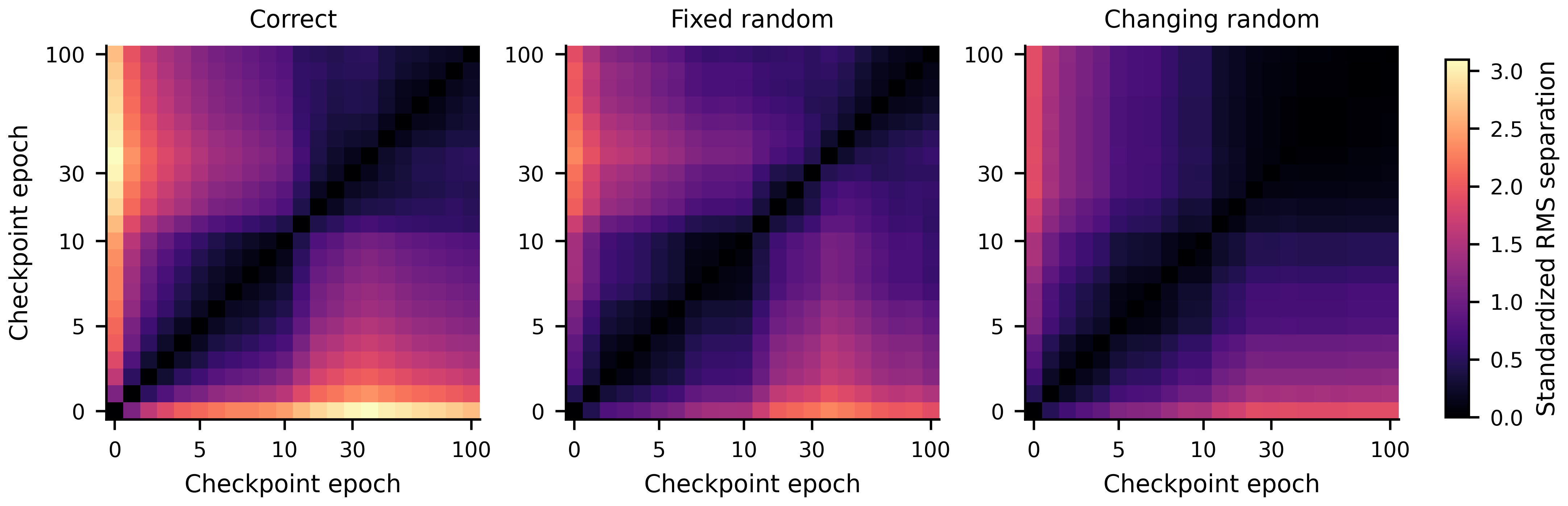}
\caption{Within-run temporal separation of six-dimensional normalized-fc3 summary states, averaged over five seeds. Features are H0 total/entropy and H1 total/entropy/maximum/count. Totals, maximum, and count use $\log(1+x)$, followed by feature-wise standardization across all displayed states. Color is Euclidean separation divided by $\sqrt{6}$. Checkpoints occupy equally spaced matrix cells despite unequal epoch intervals. These are summary-vector separations, not Wasserstein diagram distances.}
\label{fig:temporalsummary}
\end{figure*}
The symmetric matrices in Fig.~\ref{fig:temporalsummary} compare early and late states without assuming linear trajectories. Their zero diagonal is mathematical self-distance. Block structure can indicate similar summary states over a range of checkpoints, but does not establish mechanistic convergence or equal persistence diagrams: six summaries are a lossy representation. Global standardization is used only for retrospective visualization. No additional hypothesis test or classifier score is inferred from these matrices.

\section{Execution Cost, Artifact Access, and Reproduction}
\label{app:execution}
Summing recorded epoch runtimes gives 2,643.82\,s (44.06\,min) for training and end-of-epoch monitoring across the 15 runs. These timers end before persistent checkpoint writes and are not end-to-end Colab wall-clock time. Summing cached topology timers gives 191.48\,s (3.19\,min); those timers cover distance construction, Ripser, and singular-value/summary calculation, not all cloud I/O, grid encoding, matching, plots, or report export. Separate wall-clock timers for diagram matching and nested classification were not recorded and cannot be reconstructed from result values. The supplied compressed archive is 2,192,571,623 bytes (2.19\,GB decimal; approximately 2.04\,GiB).

\begin{table}[!t]
\centering
\caption{Recorded computational cost and unavailable timers.}
\label{tab:cost}
\footnotesize
\begin{tabularx}{\columnwidth}{@{}lX@{}}
\toprule
Component & Recorded quantity \\
\midrule
Training + monitoring & 2,643.82\,s; excludes checkpoint I/O \\
Cloud topology computation & 191.48\,s; excludes other stages \\
Diagram matching wall time & Not separately recorded \\
Nested prediction wall time & Not separately recorded \\
Original ZIP size & 2,192,571,623 bytes \\
Compute device & One Tesla T4; CPU PH/matching \\
\bottomrule
\end{tabularx}
\end{table}

The source package contains the user draft for traceability, official IEEEtran files, LaTeX sections/tables, 600-DPI PNG figure assets, plotting scripts, numerical CSVs, and audit records. It does not duplicate the approximately 2.19\,GB original ZIP or the model checkpoints. The companion reproduction code installs the archived modular source and explicitly retains the Wasserstein-only/deferred-bottleneck execution choice. Example commands are:
\begin{quote}\begin{minipage}{\linewidth}\footnotesize\ttfamily
python -m pip install -r requirements.txt\\
python reproduce.py --mode pilot\\
python reproduce.py --mode full\\
python reproduce.py --mode full --analysis-only
\end{minipage}\end{quote}
These commands apply to the bundled reproduction project and a full runtime artifact directory. A compact selected-weight export is not an all-epoch training backup; rebuilding every omitted checkpoint from it would require restored full weights or retraining. The manuscript's numerical tables and figures can instead be regenerated from the supplied CSVs, with the original ZIP required only for audits or displays that read cloud/diagram arrays. No additional training was performed in preparing this paper.

Known corrections are explicit: the saved schedule note says 23 instead of 22; all 282 feature-cap skips remain missing; bottleneck analysis is deferred; uncertainty intervals are unadjusted; and the classification-fold comparison is descriptive. Public release and venue-specific length/disclosure requirements remain author decisions rather than invented repository or acceptance claims.

\clearpage
\balance

\end{document}